\documentclass[preprint,12pt,authoryear]{elsarticle}

\usepackage{amssymb}
\usepackage{amsmath}
\usepackage{stfloats}
\usepackage{graphicx}
\usepackage{subcaption}
\usepackage{booktabs}
\usepackage{multirow}
\usepackage{amssymb}
\usepackage{xcolor}
\usepackage{rotating}
\usepackage{siunitx}
\usepackage[dvipsnames]{xcolor}
\usepackage{hyphenat}
\usepackage[hyphens]{url}

\usepackage{acronym} 

\usepackage[breaklinks=true]{hyperref}
\hypersetup{
     colorlinks,
     linkcolor={red!75!black},
     citecolor={blue!75!black},
     urlcolor={blue!75!black},
}

\usepackage{multicol}
\usepackage{microtype}

\newacro{AMD-HookNet}[AMD-HookNet]{attention-multihooking-deep-supervision HookNet}
\newacro{aspp}[ASPP]{Atrous Spatial Pyramid Pooling}
\newacro{caffe}[CaFFe]{``CAlving Fronts and where to Find thEm''}
\newacro{CNN}[CNN]{Convolutional Neural Network}
\newacro{crf}[CRF]{Conditional Random Field}
\newacro{dl}[DL]{Deep Learning}
\newacro{ecv}[ECV]{Essential Climate Variable}
\newacro{fov}[FoV]{forest and other vegetation}
\newacro{gee}[GEE]{Google Earth Engine}
\newacro{gla-st}[GLA-ST]{global-local attention Swin-Transformer block}
\newacro{gla}[GLA-STDeepLab]{global-local attention Swin-Transformer-based DeepLabv3+}
\newacro{gru}[GRU]{Gated Recurrent Unit}
\newacro{iou}[IoU]{Intersection over Union}
\newacro{lulc}[LULC]{Land Use and Land Cover}
\newacro{ltae}[L-TAE]{Lightweight Temporal Attention Encoder}
\newacro{mcc}[MCC]{Matthew's Correlation Coefficient}
\newacro{mde}[MDE]{Mean Distance Error}
\newacro{miou}[mIoU]{mean Intersection over Union}
\newacro{mlp}[MLP]{multilayer perceptron}
\newacro{mtl}[MTL]{multi-task learning}
\newacro{na}[NA]{``no information available''}
\newacro{oim}[ocean]{ocean and ice mélange}
\newacro{pe}[PE]{positional encoding}
\newacro{sam}[SAM]{Segment Anything Model}
\newacro{sar}[SAR]{Synthetic Aperture Radar}
\newacro{scs}[SCS]{Semantic Change Score}
\newacro{sgd}[SGD]{stochastic gradient descent}
\newacro{sits}[SITS]{Satellite Image Time Series}
\newacro{tdx}[TDX]{TanDEM-X}
\newacro{tsx}[TSX]{TerraSAR-X}
\newacro{tyriongru}[Tyrion-T-GRU]{Tyrion-T with a gated recurrent unit}
\newacro{tyrionltae}[Tyrion-T-LTAE]{Tyrion-T with a lightweight temporal attention encoder}
\newacro{tyrionconv}[Tyrion-T-Conv]{Tyrion-T with a temporal convolutional layer}
\newacro{tyrions}[Tyrion-S]{Tyrion-Small}
\newacro{tyriont}[Tyrion-T]{Tyrion-Tiny}
\newacro{ViT}[ViT]{Vision Transformer}
\newacro{wmo}[WMO]{World Meteorological Organization}
\newacro{tn}[TN]{True Negative}
\newacro{fp}[FP]{False Positive}
\newacro{fn}[FN]{False Negative}
\newacro{tp}[TP]{True Positive}

\definecolor{bleudefrance}{rgb}{0.19, 0.55, 0.91}
\definecolor{pinkx}{rgb}{1.0, 0.2, 0.7}
\definecolor{rockgray}{rgb}{0.3, 0.3, 0.3}
\definecolor{icegray}{rgb}{0.75, 0.75, 0.75}

\journal{Remote Sensing of Environment}

\begin{document}

\begin{frontmatter}

\title{Multi-temporal Calving Front Segmentation}

\author{Marcel Dreier$^\text{a}$, Nora Gourmelon$^\text{a}$, Dakota Pyles$^\text{b}$, Fei Wu$^\text{a}$, Matthias Braun$^\text{b}$, Thorsten Seehaus$^\text{b}$, Andreas Maier$^\text{a}$, Vincent Christlein$^\text{a}$} 

\affiliation{organization={Friedrich-Alexander-Universität Erlangen-Nürnberg Lehrstuhl für Informatik 5},
						addressline={Martensstr.~3}, 
            city={Erlangen},
            postcode={91058}, 
            state={Bavaria},
            country={Germany}}
\affiliation{organization={Friedrich-Alexander-Universität Erlangen-Nürnberg Institut für Geographie},
            addressline={Wetterkreuz~15}, 
            city={Erlangen},
            postcode={91058}, 
            state={Bavaria},
            country={Germany}}
\begin{abstract}
The calving fronts of marine-terminating glaciers undergo constant changes. These changes significantly affect the glacier's mass and dynamics, demanding continuous monitoring. To address this need, deep learning models were developed that can automatically delineate the calving front in Synthetic Aperture Radar imagery. However, these models often struggle to correctly classify areas affected by seasonal conditions such as ice mélange or snow-covered surfaces. To address this issue, we propose to process multiple frames from a satellite image time series of the same glacier in parallel and exchange temporal information between the corresponding feature maps to stabilize each prediction. We integrate our approach into the current state-of-the-art architecture Tyrion and accomplish a new state-of-the-art performance on the CaFFe benchmark dataset.
In particular, we achieve a Mean Distance Error of \SI{184.4}{m} and a mean Intersection over Union of $\text{83.6}$. 
\end{abstract}

%
\begin{highlights}
\item Introduction of a novel multi-temporal architecture for calving front segmentation.
\item A new state-of-the-art ensembling model for calving front segmentation.  
\item An analysis of the effects of multi-temporal strategies in the context of calving front segmentation. 
\end{highlights}


\end{frontmatter}

\section{Introduction}
\label{sec:intro}

Glaciers are particularly sensitive climate indicators~\citep{IPCC}. 
Iceberg calving at the marine-terminating glacier front is a significant mechanism of ice mass loss~\citep{Marshall.2012}. The calving front, marking the boundary between glacier and ocean, changes its position over time due to physical processes such as dynamic ice flow, calving events, and submarine melting~\citep{Benn.2014}. These spatial front shifts are important for quantifying glacier change and mass loss rates~\citep{Kochtitzky.2022_2}.
To facilitate large-scale monitoring of calving front positions, satellite imagery has become the preferred method. Since many marine-terminating glaciers are located in or near the polar regions, polar nights and extended periods of low illumination hinder the continuous acquisition of optical satellite data. \ac{sar} satellites overcome these hindrances by operating independently of sunlight. Thus, they can acquire images through cloud cover, allowing consistent monitoring of calving fronts.
However, large-scale monitoring needs vast amounts of data, making manual processing and analysis impractical. As a result, many studies focus on the automatic extraction of the calving front from \ac{sar} imagery~\citep{gourmelon2025comparison}. These models take a single image and then segment it into different zones, like ice, ocean, and rock, to later extract the calving front via post-processing.
While many approaches show promising results~\citep{Wu.2023_2,tyrion}, they often suffer from season-related artifacts~\citep{gourmelon2022calving}, such as ice mélange and snow-covered rocks, which can be difficult to distinguish from glacial ice~\citep{gourmelon2025comparison}. The problem is further amplified by the noisy nature of \ac{sar} imagery, making an accurate classification from a single image challenging.
However, satellites periodically revisit the same region, allowing for the construction of a so-called \ac{sits} over a glacier. Such time-series data hold significant potential for calving front segmentation, since seasonal features such as ice mélange and snow-covered rocks typically appear only during fleeting conditions. Multiple images might also help stabilize predictions in the presence of noise. Thus, we hypothesize that a model evaluating an entire \ac{sits} at once would inherently be more robust than a model analyzing each image individually.

One major downside of this setup comes from the additional computational cost. Models working with \ac{sits} often collapse the time series to a single prediction to avoid the cost of processing multiple images in parallel~\citep{tarasiou2023vits,garnot2021panoptic}. However, the glacier, and in particular its calving front, is typically moving between the frames of a time series, making such an approach fundamentally inaccurate. Furthermore, current architectures for calving front segmentation are already complex and computationally heavy, making it difficult to process multiple satellite images in parallel~\citep{tyrion,Wu.2023_2}. In this work, we try to address this issue by introducing a new lightweight version of the current state-of-the-art transformer architecture for calving front segmentation, Tyrion~\citep{tyrion}. Afterward, we extend the model with different temporal strategies to learn the temporal relationship between the images. To take full advantage of temporal relations, we focus on multi-temporal strategies, where we take multiple images of a \ac{sits} and compute a segmentation map for each. We achieve a new state-of-the-art performance on the \ac{caffe} benchmark dataset~\citep{gourmelon2022calving}. Our main contributions are as follows:  
\begin{enumerate}
        \item Introduction of a novel multi-temporal architecture for calving front segmentation.
    \item A new state-of-the-art ensembling model for calving front segmentation.  
    \item An analysis of the effects of multi-temporal strategies in the context of calving front segmentation. 
\end{enumerate}

The paper is structured as follows. Section~\ref{sec:relatedwork} discusses prior work in calving front segmentation and temporal processing-strategies in remote sensing. Next, Section~\ref{sec:methodology} presents our modification to the Tyrion architecture and our proposed temporal connections, which we evaluate in Section~\ref{sec:exp}. The results of these experiments are presented in Section~\ref{sec:results}, followed by an in-depth discussion in Section~\ref{sec:discussion}. Lastly, we summarize, evaluate, and conclude our work in Section~\ref{sec:conclusion}.

\section{Related Work}
\label{sec:relatedwork}

\subsection{Calving Front Delineation}
The delineation of glacier calving fronts in satellite imagery has traditionally been done manually~\citep{Baumhoer.2019}. Since 2019, a collection of studies have focused on automating this process with deep learning, introducing various methods to enhance the performance of basic deep learning models. \citet{Davari_Islam.2021, Davari_Baller.2021, gourmelon2022calving, Holzmann.2021, Mohajerani.2019} focused on mitigating the class imbalance in binary front segmentation, while some later studies~\citep{Heidler.2021, Li.2025} modeled the front directly as lines employing deep active contour models.
However, most studies perform a segmentation into landscape zones, extracting the calving front during post-processing. 
A multitude of studies~\citep{Herrmann.2023, gourmelon2022calving, Loebel.2022, Periyasamy.2022, Baumhoer.2019, Zhang.2019,ZHAO2025666} focused on optimizing the U-Net~\citep{Ronneberger.2015}, while others improve the post-processing by sorting out implausible front predictions or refining the predictions~\citep{Zhang.2021, Zhang.2023, Gourmelon.2023, Mohajerani.2019}.
More advanced techniques include multi-task learning~\citep{Cheng.2021, Heidler.2021, Herrmann.2023}, the inclusion of attention mechanisms~\citep{Heidler.2021, Zhu.2023, Holzmann.2021, Wu.2023_1, Wu.2023_2, Maslov.2023, Putatunda.2024}, the utilization of change information~\citep{ZHAO2025666}, the focus on uncertain areas to reduce uncertainty overall~\citep{Hartmann.2021}, and pretraining with large unlabeled datasets~\citep{tyrion}.
Some of the most successful studies~\citep{Wu.2023_1, Wu.2023_2, gourmelon2025comparison} invented models, with dual-branch architectures, that are able to include a large context around the calving front and process it effectively.

\citet{maslov2024glacier} introduced a deep learning framework that aggregates information from several time steps into a single zone prediction. Since this method neglects the temporal evolution of the calving front across the processed window, it would naturally be inaccurate in settings where the calving front moves between the captured frames.
To the best of our knowledge, no prior study has systematically investigated the potential of utilizing a multi-temporal deep learning model to extract the calving front from \ac{sar} imagery.

\subsection{Temporal strategies in remote sensing}

\begin{figure*}[htb]
					\centering
					\includegraphics[width=\textwidth]{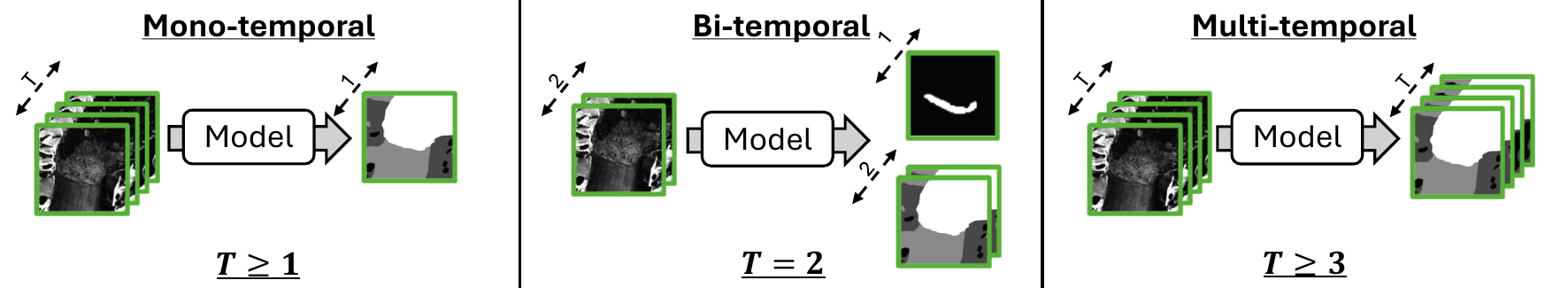}
				\caption{Overview of three temporal strategies for processing satellite image time series: (left) mono-temporal methods use one or several images and produce a single combined prediction; (middle) bi-temporal methods process pairs of images jointly to infer change or state between two time points; and (right) multi-temporal methods ingest multiple images simultaneously and generate a prediction for each time step in the series.
                }
	\label{fig:temporal_strats}
	\end{figure*}

Automatic analysis of \ac{sits} is a crucial research area with applications across a wide range of domains, including change detection, deforestation monitoring, urban planning, disaster prevention, and many more. Given the wide variety of these tasks, different temporal strategies have been employed depending on the specific task and data. We differentiate between three main categories: mono-temporal, bi-temporal, and multi-temporal (compare Fig.~\ref{fig:temporal_strats}).

~\\\noindent
\textbf{Mono-temporal} approaches take one or multiple images from the satellite time series and then make one combined prediction for all of them.~\citep{tarasiou2023vits,garnot2021panoptic,garnot2020satellite,garnot2020lightweight,ballas2015delving,shi2015convolutional}. Afterward, additional post-processing steps often derive the task-specific output. This approach offers considerable flexibility, as most state-of-the-art segmentation networks~\citep{Ronneberger.2015,chen2019rethinking,lai2021semi,liu2021swin,dosovitskiy2020image} can be used in a mono-temporal manner with no or only small modifications. However, this setup has a significant weakness regarding calving front segmentation; if the model processes each image independently to produce a single prediction, it cannot utilize any temporal information. Effectively, this makes the model non-temporal, as there is no information exchange between the different time steps~\citep{yang2022multi,cheng2023application}. This can severely limit the performance on \ac{sar} imagery in polar regions, as rocks might be temporarily covered by snow, complicating the distinction between rocky terrain and the glacier itself.
Conversely, if the model relies on multiple images to produce a single prediction, the calving front might have already moved between the frames, making a single localization not applicable for all images.

~\\\noindent
\textbf{Bi-temporal} approaches process two input images of a satellite image time series at once. They are commonly employed for change detection tasks and adopt a dual-branch architecture with shared weights. One branch processes the pre-change image, and one branch processes the post-change image, while some architectures also allow for interaction between the branches~\citep{feng2023smbcnet,marsocci2023inferring,fang2023changer,li2023lightweight,bernhard2023mapformer,zheng2022changemask,cui2023mtscd,daudt2019multitask,ding2022bi,ding2024joint,jiang2023ttnet,liu2024tbscd,tian2022large,bruzzone2002iterative,weismiller1977change,xia2022deep,yuan2022transformer,Zhao2022-ta,Zheng2022-oz}. 
The two resulting feature representations are then fused in the decoder to predict a change map~\citep{feng2023smbcnet,marsocci2023inferring,fang2023changer,li2023lightweight,daudt2019multitask}. Several methods extend this approach by incorporating semantic segmentation on each image. In that manner, the occurred change can be further described via semantic classes, similar to the mono-temporal approaches~\citep{bernhard2023mapformer,zheng2022changemask,daudt2019multitask,ding2022bi,ding2024joint,jiang2023ttnet,liu2024tbscd,tian2022large,xia2022deep,yang2020semantic,yuan2022transformer,Zhao2022-ta,Zheng2022-oz}.
One advantage of bi-temporal models is their ability to process two images from different time points simultaneously, allowing them to capture short-term temporal dynamics effectively. However, more complex or gradual temporal patterns are often challenging, as only two images do not provide enough temporal context~\citep{zhao2025histenet}. Another downside of bi-temporal models is that the dual-branch architecture is highly tailored toward change detection, limiting its application to different tasks.

~\\\noindent\textbf{Multi-temporal} approaches simultaneously process multiple images of a \ac{sits} and make a prediction for each one~\citep{he2024time,elliotvincent,saha2020unsupervised}. The temporal information flow is facilitated differently depending on the specific architecture. For example, ~\citet{elliotvincent} restructured the lightweight temporal attention encoder module~\citep{garnot2020lightweight} to enable a multi-temporal information flow, while \citet{he2024time} employed 1D convolutions to process the temporal information for every pixel individually. \citet{voelsen2024transformer} introduced two distinct branches in each layer of their model to separately process spatial and temporal information. Thereby, they can concurrently process spatial and temporal information before fusing them.
Several other approaches also employ multi-temporal feature processing, including recurrent neural networks~\citep{shi2015convolutional,ballas2015delving,papadomanolaki2019detecting,chen2024unsupervised}, temporal attention mechanisms~\citep {liu2022video,hafner2024continuous}, and 3D convolutions~\citep{tran2015learning}. 
One of the primary benefits of the multi-temporal approach lies in its broad temporal coverage. Because it can access a wider temporal context of the \ac{sits}, it can generalize to more complex temporal patterns and is more robust to outliers~\citep{arfa2024utilizing,russwurm2018multi}. These characteristics make the approach well-suited to tackle calving front segmentation in \ac{sar} imagery, as those images often suffer from seasonal variations such as ice mélange or are affected by speckle noise.
However, one downside of multi-temporal approaches is their increased computational requirements. As they must process all the images in parallel, their computational cost is almost directly proportional to the length of the time series. To mitigate this cost, we reason that a lightweight model is necessary to facilitate efficient processing.

\section{Methodology}
\label{sec:methodology}

	\begin{figure*}[t]
					\centering
					\includegraphics[width=\textwidth]{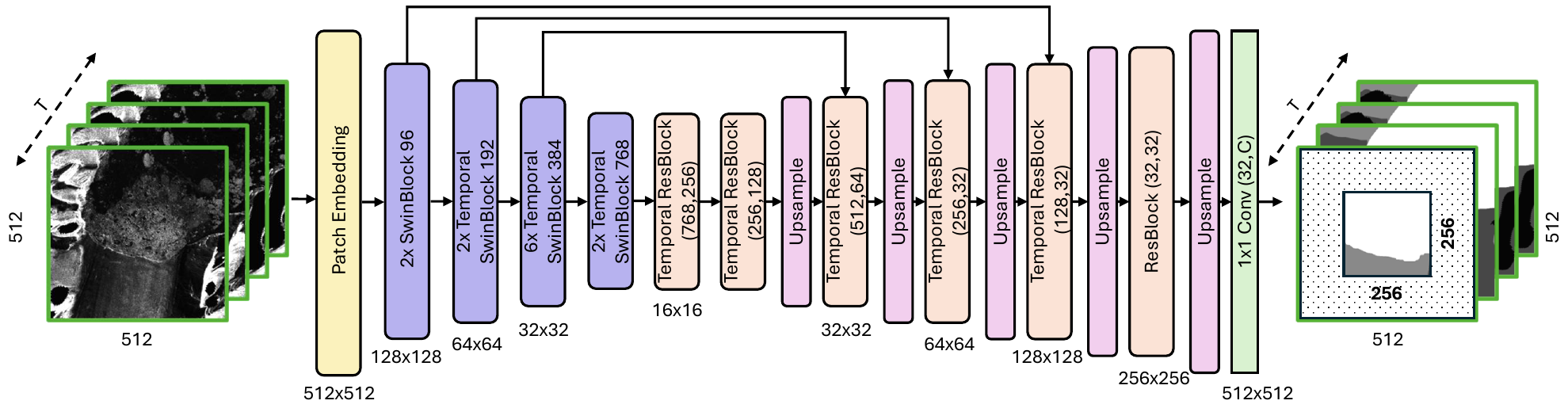}
					\caption{Overview of the proposed architecture design based on Tyrion. The structures of the temporal SwinBlock and the temporal ResBlock are depicted in Fig.~\ref{fig:swin_layer}. The numbers after each block name indicate the channel size, while the numbers below each block indicate the patch resolution. 
					The input is a \ac{sits} of length $T$ with images sized $512\times512$. While the output remains the same dimensionality, only the inner $256\times256$ pixels are used for evaluation. Input and output channels $C$ are dependent on the application. Note that for \ac{tyriont} the \ac{sits} has the length $T=1$. The illustration is based on the Figure from Gourmelon et al.~\citep{tyrion}. }
	\label{fig:tyrion_overview}
	\end{figure*}

	We design our model based on Tyrion from~\citet{tyrion}, a current state-of-the-art model for calving front segmentation. Tyrion is a U-shaped segmentation network consisting of a SwinV2 encoder~\citep{liu2022swin} and a convolutional decoder with skip connections between the two components. The SwinV2 encoder is a hierarchical Vision Transformer~\citep{dosovitskiy2020image} which utilizes a shifted window-based self-attention mechanism for feature extraction~\citep{liu2021swin}. Initially, a patch embedding layer partitions the input image into non-overlapping patches of $4 \times 4$ pixels. Afterward, these embedded patches are processed through a series of SwinBlocks with an alternating window and a shifted-window attention mechanism. Between the SwinBlocks, the feature maps are progressively downsampled with patch merging layers, enabling multi-scale representation learning. Next, the processed feature maps are forwarded into the convolutional decoder to predict a segmentation map of the image. This decoder consists of a series of ResBlocks and UpsampleBlocks based on the design of~\citet{esser2021taming}, with skip connections to preserve high-resolution spatial features. The final component of the decoder is a convolutional layer that predicts a class embedding for every pixel in the input image. 
	To incorporate a greater spatial context, it trains on patches of $512 \times 512$ pixels as input, but during inference, it only uses the inner $256 \times 256$ pixels of the prediction for evaluation. In this way, it avoids more costly two-branch architectures such as AMD-HookNet~\citep{Wu.2023_1} and HookFormer~\citep{Wu.2023_2}. 
 
	Although Tyrion performs well on mono-temporal data, its size makes it computationally expensive to extend to time series data. To mitigate this issue, we restructured its architecture, focusing on the decoder. Inspired by state-of-the-art segmentation networks with lightweight decoders~\citep{xie2021segformer, chen2019rethinking,jain2023semask,strudel2021segmenter}, we reason that we can significantly lower Tyrion's complexity by reducing the size of the decoder while retaining competitive performance.
	In particular, we reduce the decoder’s channel size in Tyrion by two-thirds and remove the skip connection in the lowest layer, as it bypasses only a single ResBlock while adding a substantial number of parameters.
	Through these modifications, we reduce Tyrion's parameter count from \SI{50.9}{M} to \SI{31.4}{M} and lower its computational complexity from \SI{162.9}{GFLOPs} to \SI{67.2}{GFLOPs}. We call this new version \acf{tyriont} and use it as a baseline to analyze the effects of the different temporal connections. Figure~\ref{fig:tyrion_overview} depicts the overall setup. To avoid confusion between the different setups, we will refer to the original Tyrion architecture as \ac{tyrions}, because its parameter count is on par with Swin-S~\citep{liu2021swin}.

	\begin{figure}[t]
					\subfloat[Temporal ResBlock]{\includegraphics[width=0.49\textwidth]{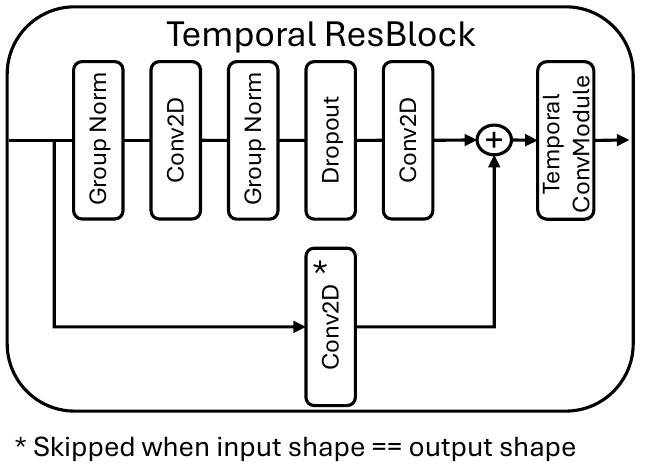}} 
					\subfloat[Temporal SwinBlock]{\includegraphics[width=0.49\textwidth]{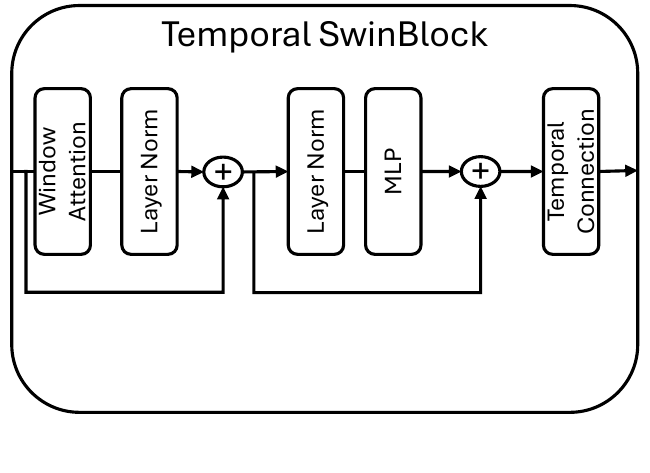}}
							\caption{Overview of our modified temporal SwinBlock and temporal ResBlock. The base designs are adapted from Liu et al.~\citep{liu2022swin} and Esser et al.~\citep{esser2021taming}, respectively. The different options for the temporal connections are depicted in Fig.~\ref{fig:temporal_connections}, including the temporal convolution layer. For the ResBlock, if the input and output channels are equal, the two-dimensional convolution layer in the skip connection is omitted.}
	\label{fig:swin_layer}
	\end{figure}

	\subsection{Temporal Information Flow}
	With \ac{tyriont} as a basis, we extend the model with temporal connections.  A straightforward approach would be to add 3D-convolutional layers~\citep{tran2015learning} or replace the SwinBlock components with their 3D counterparts~\citep{liu2022video}. However, such modifications would significantly increase the model size and computational complexity. Instead, we opt for a more efficient ``2+1'' approach, where we alternate between 2D-spatial and 1D-temporal layers, as proposed by \citet{tran2018closer}. 
	This design limits the number of additional parameters and also allows us to utilize pretrained weights from ImageNet for the 2D-SwinV2 Transformer~\citep{liu2022swin}. In detail, after every SwinBlock in the encoder and every ResBlock in the decoder, we insert a 1D temporal connection that exchanges information over the temporal axis at every point in the feature map. Thus, the network captures temporal information at different resolutions, allowing it to learn more complex temporal relationships. Figure~\ref{fig:swin_layer} illustrates the structure of the modified temporal SwinBlock and the updated temporal ResBlock. A drawback of this approach is its potential for substantial additional computational cost. To mitigate this, we restrict the temporal connections to the lowest three stages of the network and design them to remain lightweight. We also explore three different designs for the temporal connection based on established architectures to achieve a good trade-off between complexity and computation. However, we will limit the two more costly connections to the encoder to remain computationally efficient. Table~\ref{tab:tyrionComparasion} provides an overview of the different Tyrion variants and their respective costs and complexities. 

	\begin{table}[!t]
	\caption{Comparison of the number of parameters and computational complexity for the different Tyrion setups. GFLOPs are normalized to a single $256\times256$ output. Note that the input resolution is still $512\times512$.}
	\label{tab:tyrionComparasion}
	\centering
	\begin{tabular}{c|cc}
	\toprule
	Model & Parameters & FLOPs\\
	\midrule
	Tyrion-S & \SI{50.9}{M} & \SI{162.9}{G} \\
	Tyrion-T & \SI{31.4}{M} & \SI{67.2}{G} \\
	Tyrion-T-Conv & \SI{38.0}{M} & \SI{76.5}{G}\\
	Tyrion-T-LTAE & \SI{34.9}{M} & \SI{72.7}{G} \\
	Tyrion-T-GRU & \SI{41.3}{M} & \SI{71.9}{G}  \\
	\bottomrule
	\end{tabular}
	\end{table}

\begin{figure*}[t]
					\centering
					\includegraphics[width=0.85\textwidth]{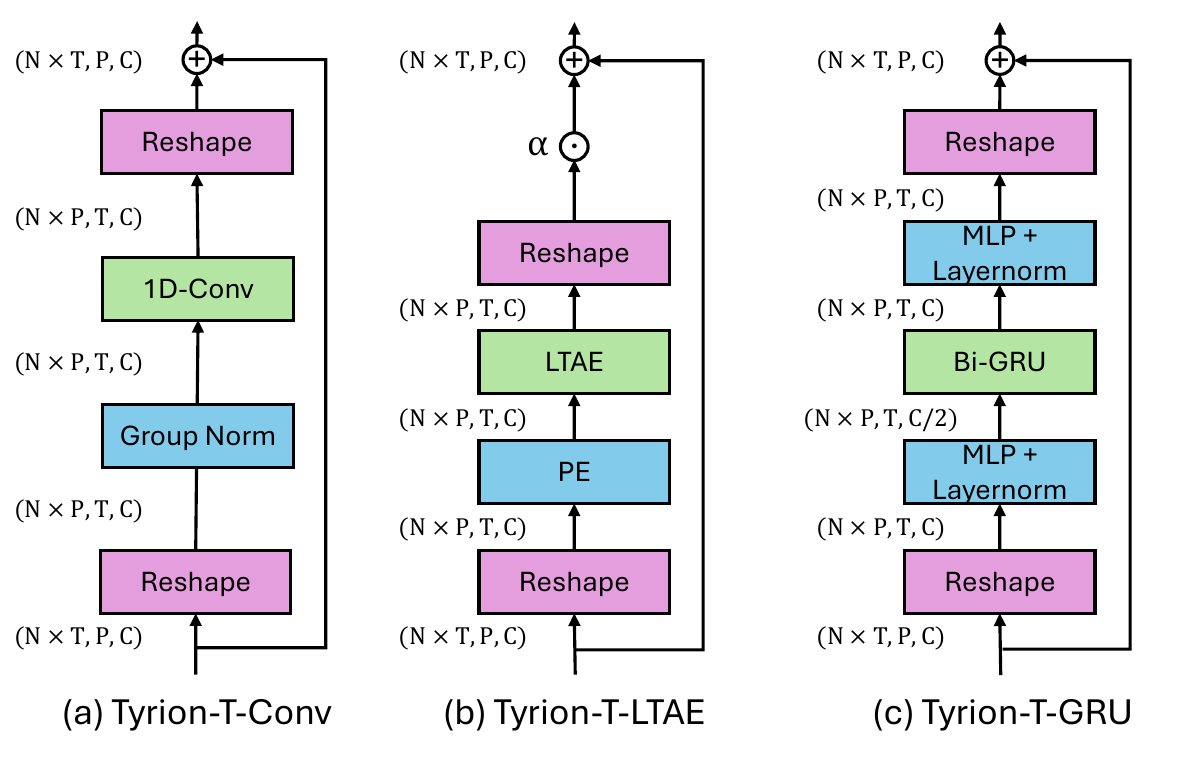}
	\caption{Overview of our proposed temporal connections: \ac{tyrionconv}, \ac{tyrionltae}, and \ac{tyriongru}. $N$ is the batch size, $T$ is the temporal sequence length, $P$ is the number of patches, $C$ is the number of channels, and $\alpha$ is a trainable weighting factor initialized as zero. MLP stands for the multi-layer perceptron, PE for positional encoding, LTAE for the lightweight attention encoder, and BI-GRU for the bidirectional gated recurrent unit. We omitted activation functions for a clearer presentation.}
	\label{fig:temporal_connections}
	\end{figure*}

	\paragraph{Tyrion-T-Conv}
	Our first and simplest design is the temporal ConvModule, inspired by \citet{tran2018closer}. This approach captures temporal patterns by applying a 1D convolution over the temporal axis. To facilitate a smooth combination of temporal and spatial features, we initialize the convolutional layer with zeros~\citep{zhang2023adding}. Additionally, we incorporate a Group Normalization Layer and an activation function to stabilize training and learn nonlinear functions~\citep{ramachandran2017searching,wu2018group}. This overall simple structure allows for fast processing while combining local temporal features. The structure of the temporal connection is depicted in Fig.~\ref{fig:temporal_connections}~(a). We call the overall Tyrion design \ac{tyrionconv}.
	\paragraph{Tyrion-T-LTAE}
	Our second approach is based on the design of the \ac{ltae} by \citet{garnot2020lightweight}. \ac{ltae} is a modified multi-head self-attention mechanism over the temporal axis of the time series, where every pixel in the input feature map attends to the same pixel at a different time step. It is very efficient due to its channel grouping strategy, where it splits the channels of the input features into groups, each being processed in parallel by a different attention head. Furthermore, \ac{ltae} reworks the classical attention mechanism of queries, keys, and values by introducing a single master query for each attention head, thus collapsing the temporal dimension. \citet{elliotvincent} replaced the single master query and instead computed a query for each token in the time series. This method preserves the temporal dimension and makes the mechanism applicable to multi-temporal applications like ours. 
	To further improve the temporal understanding, the \ac{ltae} applies 1D \ac{pe}~\citep{vaswani2017attention} based on the date of the satellite image.
	A shortcoming of the \ac{ltae} is that it is only applied once at the lowest feature resolution, limiting its capabilities in capturing multi-scale features~\citep{elliotvincent,garnot2021panoptic}. However, our encoder structure solves this issue by applying \ac{ltae} multiple times throughout the network, making the combination of \ac{tyriont} and \ac{ltae} a promising approach to explore. 
	To stabilize and accelerate the training of the network, we multiply the feature map processed by the \ac{ltae} with a trainable weight factor before summing it with the output of the skip-connection~\citep{bachlechner2021rezero}. Figure~\ref{fig:temporal_connections}~(b) depicts the structure of the proposed module, which we call \ac{tyrionltae}.

	\paragraph{Tyrion-T-GRU}
	Our third and final design is inspired by the work of Ballas et al.~\citep{ballas2015delving}, which combines spatial layers with a \ac{gru}~\citep{cho2014learning} for temporal understanding. However, \acp{gru} have a predefined one-directional information flow, limiting the temporal context in early images of the time series. Thus, we employ a bidirectional \ac{gru} that processes the sequence in both temporal directions~\citep{schuster1997bidirectional}. 
	As this would increase the computational overhead substantially, we add a \ac{mlp} before the bidirectional \ac{gru} to compress the channel size of the feature map to half its original size, thus limiting the additional computational cost. The bidirectional \ac{gru} then outputs two separate feature maps, one for each temporal direction. To combine these two feature maps, we add a final \ac{mlp} after the bidirectional \ac{gru}. Figure \ref{fig:temporal_connections}~(c) depicts the design of the temporal component, which we refer to as \ac{tyriongru}.
    
	\section{Experiments}
	\label{sec:exp}

	\subsection{Metrics}
	To assess the performance of our models on calving front segmentation, we employ the two metrics proposed by Gourmelon et al.~\citep{gourmelon2022calving}: the \ac{iou} and the \ac{mde}. The \ac{iou} measures the model's performance on the initial zone segmentation prediction $\hat{y}$ compared to its ground truth $y$~\citep{jaccard1912distribution}. It is defined for each class of the dataset as the number of \ac{tp} predictions divided by the sum of \ac{tp}, \ac{fn}, and \ac{fp} predictions. To simplify the \ac{iou} into a single value, we build the average over all the classes $C$ and refer to it as the \ac{miou}, i.\,e.:
    	\begin{multline}\label{eq:miou}
					\mathrm{mIoU}(y,\hat{y})=\frac{1}{|C|} \sum_{c \in C}\mathrm{IoU}_\text{c}(y,\hat{y}) 
					= \frac{1}{|C|} \sum_{c \in C} \frac{\mathrm{TP}_\text{c}}{\mathrm{TP}_\text{c}+\mathrm{FN}_\text{c}+\mathrm{FP}_\text{c}}
	\end{multline}
    In contrast, the \ac{mde} assesses the quality of the predicted calving front delineation $\mathcal{Q}$ in an image $\mathcal{I}$ by calculating the symmetric mean distance between the ground truth calving front $\mathcal{P}$ and $\mathcal{Q}$. It is defined as:
\begin{multline}\label{eq:mean_distance_error}
					\mathrm{MDE}(\mathcal{I}) = \frac{1}{\sum_{(\mathcal{P}, \mathcal{Q}) \in \mathcal{I}} (|\mathcal{P}| + |\mathcal{Q}|)}\\
					\cdot \sum_{(\mathcal{P}, \ \mathcal{Q}) \in I} \bigg( \sum_{\vec{p} \in \mathcal{P}} \min_{\vec{q} \in \mathcal{Q}} \lVert \vec{p}-\vec{q} \rVert_\text{2} + \sum_{\vec{q} \in \mathcal{Q}} \min_{\vec{p} \in \mathcal{P}} \lVert \vec{p}-\vec{q} \rVert_\text{2} \bigg)
	\end{multline}
	In addition to the \ac{miou} and the classical \ac{mde}, we also note the number of images where no calving front could be extracted after the post-processing as $\varnothing$. We also calculate the \ac{mde} based on the averaged labels from the multi-annotator study by \citet{gourmelon2025comparison}, denoted as $\ac{mde}_{\text{MA}}$.

	\subsection{Data}
	To assess our model, we use the \ac{caffe} benchmark dataset~\citep{gourmelon2022calving} with its official training and test splits. It contains 681 \ac{sar} images of seven marine-terminating glaciers captured by six different satellites between 1996 and 2020. As all the images are centered on the corresponding glaciers, we can define all the images of the same glacier as a time series. However, the images come in different spatial resolutions, requiring resizing to align perfectly. 
	Since this procedure could potentially introduce artifacts, we tighten the definition of the time series during evaluation to images of the same glacier with the same resolution. During training, however, we resampled the images for more diverse time series.

	For testing, we reserve all 122 satellite images of the Mapple and Columbia glaciers in the \ac{caffe} dataset. From the remaining 559 images, we randomly picked 26 samples from Jakobshaven, 13 from Jorum, and 13 from Crane for validation. This left us with 507 samples for training. Every image in the dataset has a binary label for the calving front and a separate segmentation mask, which assigns every pixel to one of four classes: \ac{na}, rock, glacier, or \ac{oim}. In addition to the official \ac{caffe} evaluation, we also evaluate our models based on the multi-annotator labels of \citet{gourmelon2025comparison} and its additional post-processing steps.

	\subsection{Experimental Protocol}
	To evaluate our modifications, we train and evaluate our three multi-temporal \ac{tyriont} setups, the reduced parameter setup \ac{tyriont}, and the current state-of-the-art model for calving front segmentation \ac{tyrions}~\citep{tyrion}. Additionally, we compare it to a more general state-of-the-art multi-temporal model proposed by Vincent et al.~\citep{elliotvincent} to see whether a multi-temporal architecture without paradigms designed for calving front segmentation can achieve competitive results. 
    
    For the mono-temporal \ac{tyrions} and \ac{tyriont}, the time series length is $T = 1$; for the multi-temporal \ac{tyriont}, $T = 8$.
    Since the \ac{caffe} dataset only covers a few images per month per glacier, a time-series length of $T=8$ still provides a wide variety of seasonal differences to improve the model's predictions. However, we chose $T=24$ for the approach by \citeauthor{elliotvincent} to remain consistent with their original experimental setup. 
	The patch size for the different Tyrion versions is $512 \times 512$, while the approach proposed by Vincent et al. uses a resolution of $128 \times 128$. Since the original images are larger than these input sizes, we first apply symmetric padding to each image and then divide them into equally sized patches before feeding them to the model.

	For training, we employ several augmentations to increase the overall variety of the data. Specifically, we utilize random horizontal and vertical flipping, random rotations, gamma adjustments, contrast adjustments, brightness adjustments, random cropping, CutMix \citep{yun2019cutmix}, and random erasure \citep{zhong2020random}. 
	Similarly to Gourmelon et al.~\citep{tyrion}, we additionally use random zooming, modified poisson noise, and the modified mixup~\citep{zhang2017mixup}. 

	For training our proposed models, we also adopt the combined dice and smoothed cross-entropy loss function from Gourmelon et al.~\citep{tyrion} and the \ac{sgd}~\citep{robbins1951stochastic,amari1993backpropagation} optimizer with a learning rate of 0.01. Additionally, if the \ac{mde} does not improve for 10 epochs, the learning rate is further reduced by a factor of 0.66.
    To provide ample time for convergence and ensure a fair comparison, we train every model for 80 epochs with 5000 time series per epoch. We keep a consistent batch size of 32 time series across all our models; however, the length of each time series varies depending on the specific model.
    We begin training the different Tyrion models using an ImageNet-pretrained~\citep {deng2009imagenet} SwinV2 encoder~\citep{liu2022swin}. 
	After training, we evaluate the checkpoint with the highest \ac{mde} for calving front segmentation. We repeat every setup five times and present their mean and variance to determine the statistical error. To show the full potential of our proposed approach, we also include an ensemble setup for our proposed models over the five conducted runs. Section~\ref{appendix:hyper} gives a more in-depth overview of each model configuration.

	\subsection{Post-Processing}
	We train our model to assign a semantic class to every pixel of the input. From these zone predictions, we extract the calving fronts by following the post-processing steps of \citet{gourmelon2022calving}. These include finding the largest cluster of \ac{oim} predictions, filling any gaps inside the cluster, and then extracting the border between the \ac{oim} and glacier class as the calving front. To avoid false-positive predictions, any calving front shorter than \SI{750}{m} is deleted. For the comparison with the multi-annotator study, we also add a static rock mask to the predictions and delete any resulting fronts shorter than \SI{750}{m}, mimicking the post-processing steps of~\citet{gourmelon2025comparison}.

	\section{Results}
	\label{sec:results}

    \begin{table*}[htbp]
			\centering
				 \begin{minipage}{\textwidth}
                 \centering
                 \small
					 \noindent
					\caption{This table summarizes our evaluation of the calving front segmentation on the \ac{caffe} dataset. We train each architecture five times and present the mean and standard deviation of the results. Bold values indicate the best performance in their respective category. The \ac{mde} and \ac{mde}$_{\text{MA}}$ is presented in meters, and $\varnothing$ stands for the number of images without a detected calving front. The ensemble runs are a combination of the five conducted runs.}
					\setlength{\tabcolsep}{.55em}
					\label{tab:distance_errors_annotators}
				\begin{tabular}{c|ccc}
							\toprule
							& \multicolumn{3}{c}{\textit{Calving Front Segmentation}}  \\
							 Model & MDE $\downarrow$ & $\ac{mde}_{\text{MA}}$ $\downarrow$ & $\varnothing \in 122 \downarrow$  \\
							\midrule
							 Vincent et al.  & $850.2\pm179.3$ & $871.5\pm220.5$ & $15\pm6.5$ \\
							 Tyrion  & $306\pm33.5$ & $129.9\pm17.1$ & $0.2\pm0.4$   \\
							\midrule
								\ac{tyriont}  & $317.4\pm26.7$ & $143.9\pm25.5$ & \textbf{0.0} \\
							 \ac{tyrionconv}  & $247.6\pm17.0$ & \textbf{78.2$\pm$10.4} & \textbf{0.0}   \\
								\ac{tyrionltae} &  $232.9\pm7.5$ & $92.4\pm6.4$ & \textbf{0.0}  \\
								\ac{tyriongru} & $\textbf{202.7}\pm\textbf{27.6}$ & $88.1\pm16.0$ & \textbf{0.0}  \\
								\midrule
								\multicolumn{4}{c}{\textit{Ensembling}} \\
									\ac{tyriont}  & $296.3$ & $124.6$ & \textbf{0.0} \\
									\ac{tyrionconv} & $231.2$ & $\textbf{72.8}$ & \textbf{0.0}   \\
									\ac{tyrionltae}  & $220.4$ & $84.5$ & \textbf{0.0}    \\
									\ac{tyriongru} & $\textbf{184.4}$ & $76.5$ & \textbf{0.0}   \\
							\bottomrule
					\end{tabular}
					 \end{minipage}
	\end{table*}

    \begin{table*}[htbp]
			\centering
				 \begin{minipage}{\textwidth}
                 \centering
					 \small	
					 \noindent
					\caption{Summary of the evaluation results on the zone segmentation task on the \ac{caffe} dataset. Each architecture was trained five times; the mean and standard deviation are reported. Bold values indicate the best performance in their respective category. The ensemble runs are a combination of the five conducted runs.}
					\setlength{\tabcolsep}{.55em}
					\label{tab:distance_errors_annotators2}
				\begin{tabular}{c|ccccc}
							\toprule
							& \multicolumn{4}{c}{\textit{Zone Segmentation \ac{iou}}} \\
							 Model &  All$\uparrow$ & \ac{na}$\uparrow$  & Rock$\uparrow$  &  Glacier$\uparrow$ & Ocean$\uparrow$  \\
							\midrule
							 Vincent et al.  & $56.1\pm3.4$ & $84.8\pm3.8$ & $43.9\pm3.0$ & $56.5\pm3.0$ & $39.2\pm11.9$  \\
							 Tyrion  & $77.9\pm0.6$ & $93.6\pm0.5$ & $59.2\pm2.6$ & $73.1\pm0.4$ & $85.5\pm3.7$  \\
							\midrule
								\ac{tyriont}  & $78.7\pm0.8$ & $93.6\pm0.8$ & $58.4\pm1.4$ & $73.7\pm0.7$ & $88.9\pm0.7$  \\
							 \ac{tyrionconv}  & $81.2\pm0.4$ & $93.8\pm0.8$ & $63.2\pm1.1$ & $76.2\pm0.4$ & $\textbf{91.6}\pm\textbf{0.3}$  \\
								\ac{tyrionltae} & $81.8\pm0.9$ & $94.6\pm0.4$ & $64.9\pm 2.5$& $76.3\pm0.8$ & $91.5\pm0.4$   \\
								\ac{tyriongru} & $\textbf{82.1}\pm\textbf{0.8}$ & $\textbf{95.1}\pm\textbf{0.3}$ & $\textbf{65.3}\pm\textbf{2.0}$ & $\textbf{76.5}\pm\textbf{0.9}$ & $\textbf{91.6}\pm\textbf{0.4}$  \\
								\midrule
								\multicolumn{6}{c}{\textit{Ensembling}} \\
									\ac{tyriont}  & $79.8$ & $94.2$ & $59.3$ & $74.7$ & $91.0$  \\
									\ac{tyrionconv} & $82.1$ & $94.3$ & $64.7$ & $77.1$ & $92.4$  \\
									\ac{tyrionltae}  & $83.1$ & $95.0$ & $67.3$ & $77.7$ & $\textbf{92.6}$  \\
									\ac{tyriongru} & $\textbf{83.6}$ & $\textbf{95.5}$ & $\textbf{68.3}$ & $\textbf{78.3}$ & $92.3$  \\
							\bottomrule
					\end{tabular}
					 \end{minipage}
	\end{table*}

	\begin{figure*}[htbp]
	\includegraphics[width=\textwidth]{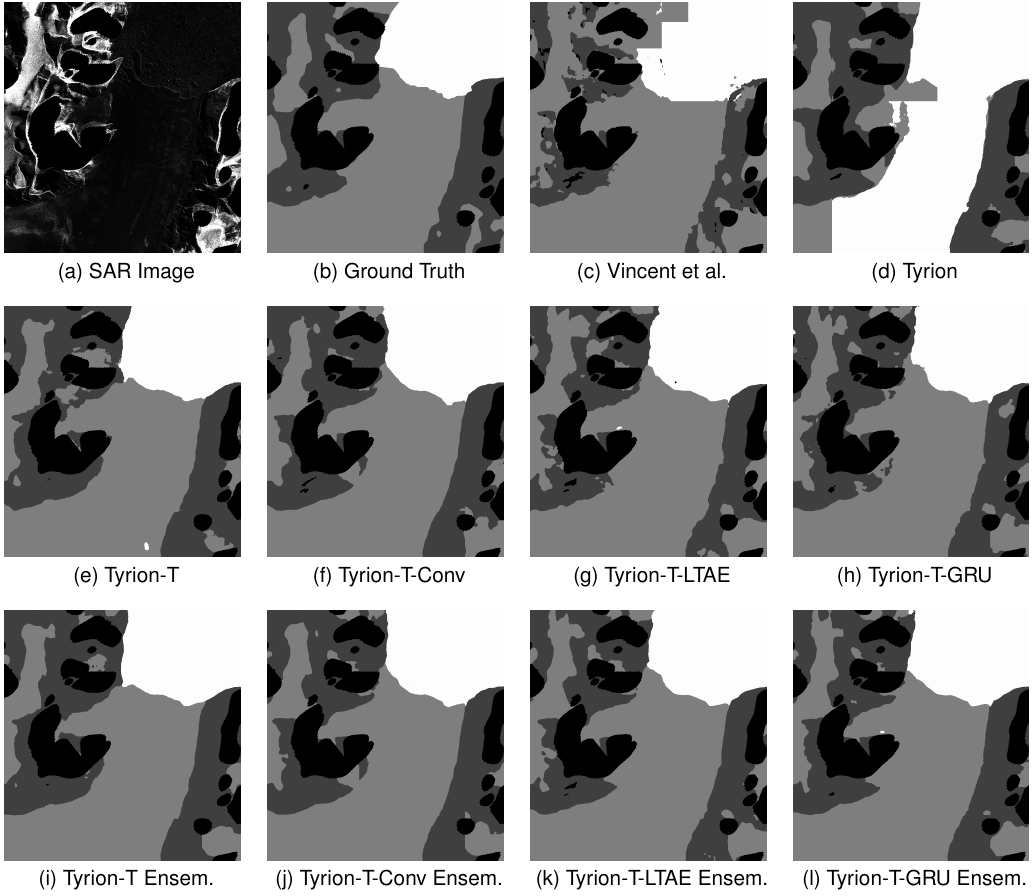}
	\caption{Qualitative comparison of the different approaches on a sample from the Mapple Glacier on the 5th of November 2010 captured by the TerraSAR-X satellite. The \ac{oim} class is \framebox{white}, the glacier class is \fboxsep=1pt\colorbox{icegray!100}{light gray}, the rock class is \fboxsep=1pt\colorbox{rockgray!100}{\textcolor{white}{dark gray}}, and the \ac{na} class is \fboxsep=1pt\colorbox{black!100}{\textcolor{white}{black}}.}
	\label{quality_maple}
	\end{figure*}

	Table~\ref{tab:distance_errors_annotators} and Table~\ref{tab:distance_errors_annotators2} summarize our results. The results show that \ac{tyriont} performs comparably to \ac{tyrions}. Furthermore, each proposed multi-temporal \ac{tyriont} setup substantially outperforms the mono-temporal versions across our recorded metrics. Within the multi-temporal setups, the \ac{tyriongru} achieves the lowest \ac{mde} and hightest \ac{miou}, followed by \ac{tyrionltae} and \ac{tyrionconv}. When focusing on the zone segmentation task, all three multi-temporal \ac{tyriont} setups show similar performance for the \ac{oim}, \ac{na}, and glacier classes. The main difference lies in the rock class, where \ac{tyrionconv} falls slightly behind. A visual comparison of the predicted segmentation masks in Fig.~\ref{quality_maple} reveals minimal qualitative differences between the multi-temporal \ac{tyriont} models. However, the mono-temporal Tyrion setups appear more prone to outliers. Especially, \ac{tyrions} over-predicts the \ac{oim} class considerably. These results are also reflected in the quantitative analysis, as the \ac{miou}s of the two mono-temporal Tyrion setups are two to four percentage points lower than the ones of the multi-temporal configurations. Interestingly, the approach from \citet{elliotvincent} has the lowest \ac{miou} of all compared models despite incorporating temporal connections. For the calving front predictions, it also has the highest \ac{mde} and a substantial number of missing fronts compared to the Tyrion setups. 

\begin{figure*}[!htbp]
	\centering
	\includegraphics[width=0.9\textwidth]{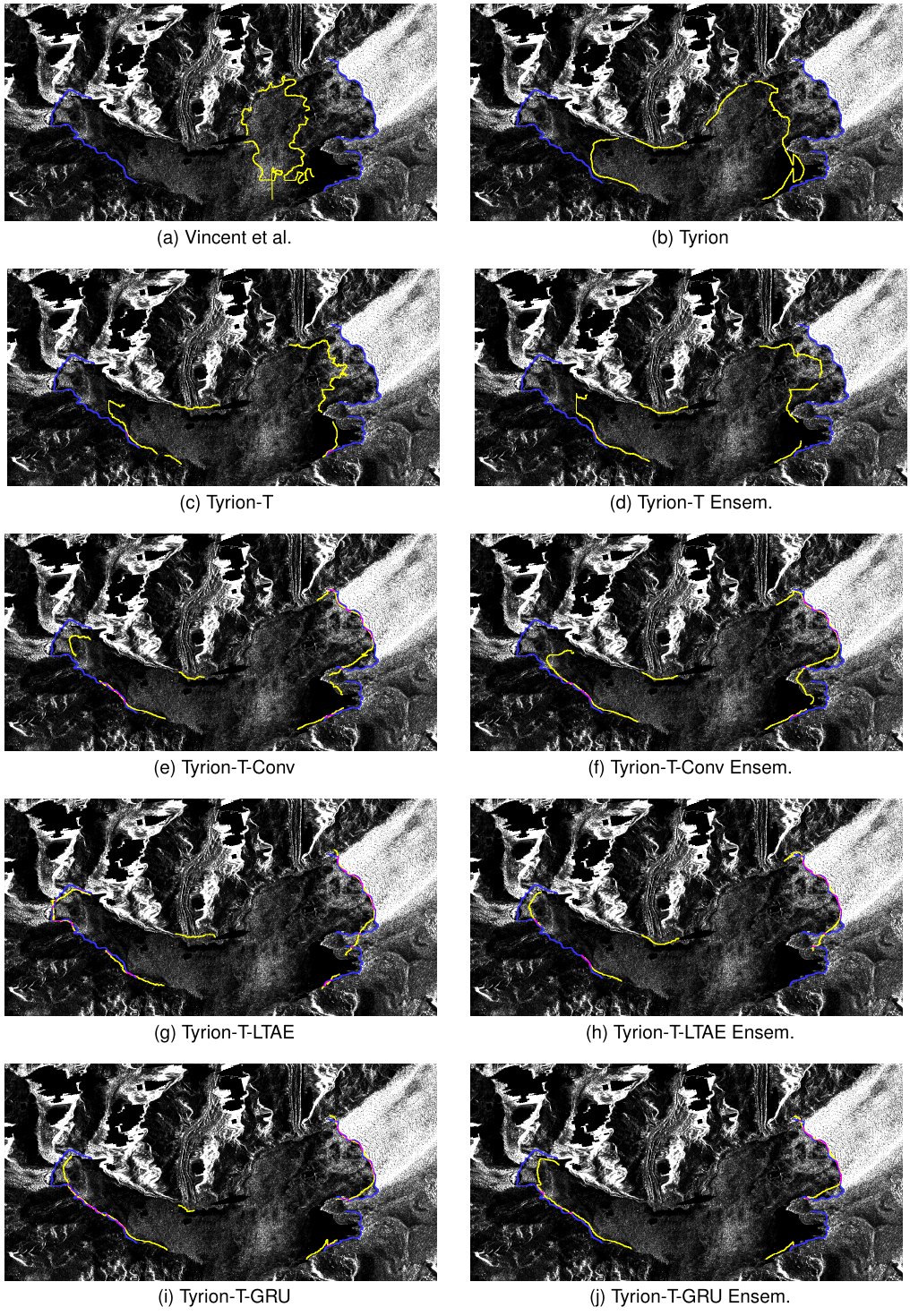}
	\caption{Qualitative comparison of the different approaches. The \ac{sar} image shows the Columbia Glacier on the 8th of March 2020 captured by the Sentinel-1 satellite. The ground truth calving front is annotated in \fboxsep=1pt\colorbox{bleudefrance!100}{blue}, the predicted calving front in \fboxsep=1pt\colorbox{yellow!100}{yellow}, and overlaps in \fboxsep=1pt\colorbox{pinkx!100}{pink}.}
	\label{calving_fronts:quality}
	\end{figure*}

	When comparing the calving front predictions of the different Tyrion setups, we observe the multi-temporal approaches outperforming the mono-temporal setups substantially. Their differences become even more apparent when visually comparing the extracted calving fronts, as illustrated by an example in Fig.~\ref{calving_fronts:quality}. The mono-temporal Tyrion setups demonstrate substantial difficulty in distinguishing between ocean and ice mélange and glacier ice; this considerably shifts the calving front towards the sea. In contrast, the multi-temporal \ac{tyriont} versions perform better, as only minor inaccuracies near the calving front appear. These inaccuracies are substantially smaller in extent than an entire ice mélange field, leading to a decreased error. 

	Lastly, when taking a closer look at the ensembling approaches for our proposed \ac{tyriont} setups, we observe a slight improvement in the overall performance for each setup. In particular, \ac{tyriongru} achieves a new state-of-the-art performance with an \ac{mde} of \SI{184.4}{m} and an \ac{miou} of 83.6.

	\section{Discussion and Outlook}
	\label{sec:discussion}
	Compared to the mono-temporal setups, our three proposed multi-temporal \ac{tyriont} setups lead to substantial improvements. This was particularly evident in areas that changed slowly or not at all, such as the rock class, which saw the most substantial gains. The additional temporal context also proved beneficial in ambiguous cases where ice mélange covered parts of the ocean. The mono-temporal model would often confuse these areas with glacial ice. We attribute this improvement to the temporal context: As the model can see the same region at multiple timesteps, we hypothesize that the model can use the information from timesteps, where it can clearly detect the ocean, to help the prediction in ambiguous cases where it sees a large amount of ice mélange. Thereby, the model can more easily distinguish between the temporary ice mélange and the glacier ice. However, the ice mélange area close to the calving front remains a challenge, as the glacier calving in this zone. In the early time steps, larger icebergs may still be present in these areas and are very close to glacier tongue, whereas in later timesteps, these icebergs may have drifted away from the glacier front or disintegrated in smaller parts and joined the ice mélange. Thus, the model still struggles to distinguish between glacial ice, freshly claved of icebergs and ice mélange in this limited area.

    Between our three proposed multi-temporal setups, we observed slight performance differences in calving front segmentation. \ac{tyriongru} performed generally the best, closely followed by \ac{tyrionltae} and Tyrion\--T\--Conv. These performance differences are reflected in the complexity of the temporal connections, as \ac{tyriongru} is the most complex and  \ac{tyrionconv} the least complex. This leads us to the assumption that the structure of the temporal connections plays a pivotal role in the overall performance of the system and should be further explored in future research. Methods that can encode the acquisition time have the potential to deal with irregularities in the time differences, and thus might be better suited to capture fast-changing areas near the calving front. 
    Interestingly, when we evaluate the models with the annotations from the multi-annotator study~\citep{gourmelon2025comparison}, the errors become considerably smaller with \ac{tyrionconv} taking the lead for the lowest \ac{mde}$_{\text{MA}}$. We attribute this shift to the expanded post-processing by \citet{gourmelon2025comparison}, which adds a static rock outcrop mask as lateral boundary of the calving front before extracting the calving fronts. 
    
    From our results, we can also see that the architectural paradigms designed for calving front segmentation have a larger impact on the results than the multi-temporal structure. The multi-temporal model from ~\citet{elliotvincent} struggled far more than any of the proposed Tyrion versions. This result highlights the need for specialized multi-temporal models for calving front segmentations, as generic solutions might fail to capture the complex nature of the \ac{sar} imagery.  

	\section{Conclusion}
	\label{sec:conclusion}

	In this study, we introduce a novel multi-temporal model designed for calving front segmentation. Our approach builds upon the Tyrion architecture proposed by Gourmelon et al.~\citep{tyrion} and incorporates several modifications, such as a smaller decoder and the integration of temporal connections. To avoid heavy computational cost, we also divided the spatial and temporal processing into separate stages.
	For the temporal connections, we implemented and tested three different designs, which we refer to as \ac{tyrionconv}, \ac{tyrionltae}, and \ac{tyriongru}. Among these three, \ac{tyriongru} demonstrated the best performance, achieving state-of-the-art results for calving front segmentation.
	Specifically, we achieved a new state-of-the-art with an \ac{mde} of \SI{184.4}{m}, and an \ac{miou} of $83.6$ on the \ac{caffe} dataset. When compared with the annotations from multiple annotators, we achieved a new state-of-the-art $\ac{mde}_{\text{MA}}$ of \SI{72.2}{m}, narrowing the gap to the average human error of \SI{38}{m}.

	\appendix
	\section{Hyperparameters}
	\label{appendix:hyper}
	This section gives an in-depth overview of the different model configurations and setups. The chosen hyperparameters are summarized in Table~\ref{tab:hyper_cfs}. To increase the variety of the data, we also employ several augmentations. We apply rotations and horizontal/vertical flips with a probability of 0.5. If we flip or rotate a single image, we must flip every image in the time series so the samples remain spatially aligned. However, for augmentations such as brightness, contrast, or gamma correction, we chose to apply the augmentation on a per-image basis with a probability of 0.2, so it would more closely resemble cases where we had images from different sensors. Additionally, we employ MixUp, CutMix, and Random Erasure with a probability of 0.1, and introduce random noise or adjust image resolution with probabilities of 0.2 and 0.3, respectively.

	\begin{table}[!htbp]
    \centering
			\begin{minipage}{\textwidth}
			\centering
			\caption{This table summarizes our model configuration. $\epsilon$ is the smoothing factor of the smoothed cross-entropy loss (s. CE). The channel dimension of each stage in the network is scaled according to the channels and the corresponding channel\_mult value. Tyrion's scheduler reduces the learning rate on a plateau (Rop).
	}
			\label{tab:hyper_cfs}
			\scriptsize

				\begin{tabular}{c|cccc}
							\toprule
								& \ac{tyriont} & \ac{tyrionconv}  & \ac{tyrionltae} & \ac{tyriongru}   \\
							 \midrule
							 Channels  & 96 & 96 & 96 & 96  \\
							Channel\_mult & [1,2,4,8] & [1,2,4,8]  & [1,2,4,8] & [1,2,4,8] \\
							\midrule
							Context\_size & $\text{512}\times\text{512}$ & $\text{512}\times\text{512}$ & $\text{512}\times\text{512}$ & $\text{512}\times\text{512}$  \\
							Patch\_size & $\text{256}\times\text{256}$ & $\text{256}\times\text{256}$ & $\text{256}\times\text{256}$ & $\text{256}\times\text{256}$  \\
							Temporal length & 1 & 8 & 8 & 8  \\
							\midrule
							 Loss & s. CE + Dice & s. CE + Dice & s. CE + Dice & s. CE + Dice  \\
							 $\epsilon$ & 0.1 & 0.1 & 0.1 & 0.1  \\
							 Optimizer & SGD & SGD & SGD & SGD  \\ 
							 Learning Rate & 0.01 & 0.01 & 0.01 & 0.01 \\
							 Scheduler & RoP  & RoP & RoP & RoP  \\
							 Warm-up steps & - & - & - & -  \\
							\midrule
							Parameters & 31.4M  & 38.0M & 34.9M  & 41.3M \\
							Flops\footnote{normalized to a single $256\times256$ patch and measured in GFLOPs} & 67.2  & 76.5 & 72.7 &  71.9 \\

							\bottomrule
					\end{tabular}
			\end{minipage}
	\end{table}

	\begin{table}[!htbp]
    \centering
			\begin{minipage}{\textwidth}
			\centering
			\caption{This table summarizes the model configuration of the comparison methods. The AdamW optimizer is based on the work from \citet{loshchilov2017decoupled}. $\epsilon$ is the smoothing factor of the smoothed cross-entropy loss (s. CE). The channel dimension of each stage in the network is scaled according to the channels and the corresponding channel\_mult value. Tyrion's scheduler reduces the learning rate on a plateau (Rop).
	}
			\label{tab:hyper_cfs}
			\scriptsize

				\begin{tabular}{c|cc}
							\toprule
								& \ac{tyrions} & Vincent et al.  \\
							 \midrule
							 Channels & 96 & 512 \\
							Channel\_mult & [1,2,4,8] & [$\frac{1}{2}$,$\frac{1}{2}$,$\frac{1}{2}$,1]\\
							\midrule
							Context\_size & $\text{512}\times\text{512}$ & - \\
							Patch\_size & $\text{256}\times\text{256}$  & $\text{128}\times\text{128}$ \\
							Temporal length & 1 & 24 \\
							\midrule
							 Loss & s. CE + Dice & CE \\
							 $\epsilon$ & 0.1 & 0.0 \\
							 Optimizer & SGD & AdamW \\ 
							 Learning Rate &  0.01 & $\text{10}^{-4}$ \\
							 Scheduler &  RoP & Constant \\
							 Warm-up steps & - & 500 \\
							\midrule
							Parameters & 50.9M  & 16.2M \\
							Flops\footnote{normalized to a single $256\times256$ patch and measured in GFLOPs} & 162.9 & 516.8 \\

							\bottomrule
					\end{tabular}
			\end{minipage}
	\end{table}

\section*{AI Declaration Statement}
During the preparation of this work, AI technologies were used to assist in the writing process. Specifically, Grammarly (Grammarly, Inc., San Francisco, CA, USA) was used in order to check for grammar and style consistency and DeepL (DeepL SE, Cologne, Germany) and ChatGPT (GPT‐4) (OpenAI, San Francisco, CA, USA) were used in order to assist with rephrasing and improving readability. After using these tools, the manuscript was carefully reviewed and the content was edited as needed. No tools or services were used for content generation.

\section*{Code and Data Availability}
We will make the code publicly available on GitHub at \url{https://github.com/ki7077/Multi-Temporal-Tyrion} after acceptance. The CaFFe benchmark dataset is already publicly available at \url{https://doi.pangaea.de/10.1594/PANGAEA.940950}.

\section*{Acknowledgments}
This research was funded by the Bayerisches Staatsministerium für Wissenschaft und Kunst within the Elite Network Bavaria with the Int. Doct. Program ``Measuring and Modelling Mountain Glaciers in a Changing Climate'' (IDP M3OCCA) as well as the German Research Foundation (DFG) project ``Large-scale Automatic Calving Front Segmentation and Frontal Ablation Analysis of Arctic Glaciers using Synthetic-Aperture Radar Image Sequences (LASSI)'' (Project number: 512625584), and the project ``PAGE'' within the DFG Emmy-Noether-Programme (DFG – SE3091/3-1; DFG – CH2080/5-1; DFG – SE3091/4-1).
The authors gratefully acknowledge the scientific support and HPC resources provided by the Erlangen National High Performance Computing Center (NHR@FAU) of the Friedrich-Alexander-Universität Erlangen-Nürnberg (FAU) under the NHR projects b110dc and b194dc. NHR funding is provided by federal and Bavarian state authorities. NHR@FAU hardware is partially funded by the DFG – 440719683.
The author team acknowledges the provision of satellite data under various AOs from respective space agencies (DLR, ESA, JAXA, CSA).

\section*{Declaration of competing interest}
The authors declare that they have no known competing financial interests or personal relationships that could have appeared to influence the work reported in this paper.

\section*{Author contribution}
\noindent
\textbf{Marcel Dreier}: Conceptualization, Methodology, Software, Project administration, Writing - Original draft preparation.
\textbf{Nora Gourmelon}: Methodology, Software, Writing - review \& editing.  
\textbf{Dakota Pyles}: Writing - review \& editing. 
\textbf{Fei Wu}: Writing - review \& editing. 
\textbf{Thorsten Seehaus}: Supervision, Writing - review \& editing. \textbf{Matthias Braun}: Supervision, Writing – review \& editing. \textbf{Andreas Maier}: Supervision, Writing – review \& editing. \textbf{Vincent Christlein}: Supervision, Writing - review \& editing.

\bibliographystyle{elsarticle-harv} 
\bibliography{references}

@Article{gourmelon2022calving,
AUTHOR = {Gourmelon, N. and Seehaus, T. and Braun, M. and Maier, A. and Christlein, V.},
TITLE = {Calving fronts and where to find them: a benchmark dataset and methodology for automatic glacier calving front extraction from synthetic aperture radar imagery},
JOURNAL = {Earth System Science Data},
VOLUME = {14},
YEAR = {2022},
NUMBER = {9},
PAGES = {4287--4313},
URL = {https://essd.copernicus.org/articles/14/4287/2022/},
DOI = {10.5194/essd-14-4287-2022}
}

@article{elliotvincent,
  publtype={informal},
  author={Elliot Vincent and Jean Ponce and Mathieu Aubry},
  title={Satellite Image Time Series Semantic Change Detection: Novel Architecture and Analysis of Domain Shift},
  year={2024},
  cdate={1704067200000},
  journal={CoRR},
  volume={abs/2407.07616},
  url={https://doi.org/10.48550/arXiv.2407.07616}
}

@article{jaccard1912distribution,
author = {Jaccard, Paul},
title = {THE DISTRIBUTION OF THE FLORA IN THE ALPINE ZONE.},
journal = {New Phytologist},
volume = {11},
number = {2},
pages = {37-50},
doi = {https://doi.org/10.1111/j.1469-8137.1912.tb05611.x},
url = {https://nph.onlinelibrary.wiley.com/doi/abs/10.1111/j.1469-8137.1912.tb05611.x},
year = {1912}
}

@InProceedings{garnot2020lightweight,
author="Garnot, Vivien Sainte Fare
and Landrieu, Loic",
editor="Lemaire, Vincent
and Malinowski, Simon
and Bagnall, Anthony
and Guyet, Thomas
and Tavenard, Romain
and Ifrim, Georgiana",
title="Lightweight Temporal Self-attention for Classifying Satellite Images Time Series",
booktitle="Advanced Analytics and Learning on Temporal Data",
year="2020",
publisher="Springer International Publishing",
address="Cham",
pages="171--181",
isbn="978-3-030-65742-0"
}

@INPROCEEDINGS{liu2022video,
  author={Liu, Ze and Ning, Jia and Cao, Yue and Wei, Yixuan and Zhang, Zheng and Lin, Stephen and Hu, Han},
  booktitle={2022 IEEE/CVF Conference on Computer Vision and Pattern Recognition (CVPR)}, 
  title={Video Swin Transformer}, 
  year={2022},
  volume={},
  number={},
  pages={3192-3201},
  doi={10.1109/CVPR52688.2022.00320}}

@INPROCEEDINGS {tran2015learning,
author = { Tran, Du and Bourdev, Lubomir and Fergus, Rob and Torresani, Lorenzo and Paluri, Manohar },
booktitle = { 2015 IEEE International Conference on Computer Vision (ICCV) },
title = {{ Learning Spatiotemporal Features with 3D Convolutional Networks }},
year = {2015},
volume = {},
ISSN = {2380-7504},
pages = {4489-4497},
doi = {10.1109/ICCV.2015.510},
url = {https://doi.ieeecomputersociety.org/10.1109/ICCV.2015.510},
publisher = {IEEE Computer Society},
address = {Los Alamitos, CA, USA},
month =Dec}

@INPROCEEDINGS{tran2018closer,
author={Tran, Du and Wang, Heng and Torresani, Lorenzo and Ray, Jamie and LeCun, Yann and Paluri, Manohar},
  booktitle={2018 IEEE/CVF Conference on Computer Vision and Pattern Recognition}, 
  title={A Closer Look at Spatiotemporal Convolutions for Action Recognition}, 
  year={2018},
  volume={},
  number={},
  pages={6450-6459},
  doi={10.1109/CVPR.2018.00675}
}

@article{shi2015convolutional,
  title={Convolutional LSTM network: A machine learning approach for precipitation nowcasting},
  author={Shi, Xingjian and Chen, Zhourong and Wang, Hao and Yeung, Dit-Yan and Wong, Wai-Kin and Woo, Wang-chun},
  journal={Advances in neural information processing systems},
  volume={28},
  year={2015}
}

@inproceedings{cho2014learning,
    title = "Learning Phrase Representations using {RNN} Encoder{--}Decoder for Statistical Machine Translation",
    author = {Cho, Kyunghyun  and
      van Merri{\"e}nboer, Bart  and
      Gulcehre, Caglar  and
      Bahdanau, Dzmitry  and
      Bougares, Fethi  and
      Schwenk, Holger  and
      Bengio, Yoshua},
    editor = "Moschitti, Alessandro  and
      Pang, Bo  and
      Daelemans, Walter",
    booktitle = "Proceedings of the 2014 Conference on Empirical Methods in Natural Language Processing ({EMNLP})",
    month = oct,
    year = "2014",
    address = "Doha, Qatar",
    publisher = "Association for Computational Linguistics",
    url = "https://aclanthology.org/D14-1179/",
    doi = "10.3115/v1/D14-1179",
    pages = "1724--1734"
}

@inproceedings{vaswani2017attention,
 author = {Vaswani, Ashish and Shazeer, Noam and Parmar, Niki and Uszkoreit, Jakob and Jones, Llion and Gomez, Aidan N and Kaiser, \L ukasz and Polosukhin, Illia},
 booktitle = {Advances in Neural Information Processing Systems},
 editor = {I. Guyon and U. Von Luxburg and S. Bengio and H. Wallach and R. Fergus and S. Vishwanathan and R. Garnett},
 pages = {},
 publisher = {Curran Associates, Inc.},
 title = {Attention is All you Need},
 url = {https://proceedings.neurips.cc/paper_files/paper/2017/file/3f5ee243547dee91fbd053c1c4a845aa-Paper.pdf},
 volume = {30},
 year = {2017}
}

@article{gourmelon2025comparison,
  title={Comparison Study: Glacier Calving Front Delineation in Synthetic Aperture Radar Images With Deep Learning},
  author={Gourmelon, Nora and Heidler, Konrad and Loebel, Erik and Cheng, Daniel and Klink, Julian and Dong, Anda and Wu, Fei and Maul, Noah and Koch, Moritz and Dreier, Marcel and others},
  journal={arXiv preprint arXiv:2501.05281},
  year={2025}
}

@INPROCEEDINGS{deng2009imagenet,
  author={Deng, Jia and Dong, Wei and Socher, Richard and Li, Li-Jia and Kai Li and Li Fei-Fei},
  booktitle={2009 IEEE Conference on Computer Vision and Pattern Recognition}, 
  title={ImageNet: A large-scale hierarchical image database}, 
  year={2009},
  volume={},
  number={},
  pages={248-255},
  doi={10.1109/CVPR.2009.5206848}
}

@INPROCEEDINGS{liu2022swin,
  author={Liu, Ze and Hu, Han and Lin, Yutong and Yao, Zhuliang and Xie, Zhenda and Wei, Yixuan and Ning, Jia and Cao, Yue and Zhang, Zheng and Dong, Li and Wei, Furu and Guo, Baining},
  booktitle={2022 IEEE/CVF Conference on Computer Vision and Pattern Recognition (CVPR)}, 
  title={Swin Transformer V2: Scaling Up Capacity and Resolution}, 
  year={2022},
  volume={},
  number={},
  pages={11999-12009},
  doi={10.1109/CVPR52688.2022.01170}
}

@INPROCEEDINGS{yun2019cutmix,
  author={Yun, Sangdoo and Han, Dongyoon and Chun, Sanghyuk and Oh, Seong Joon and Yoo, Youngjoon and Choe, Junsuk},
  booktitle={2019 IEEE/CVF International Conference on Computer Vision (ICCV)}, 
  title={CutMix: Regularization Strategy to Train Strong Classifiers With Localizable Features}, 
  year={2019},
  volume={},
  number={},
  pages={6022-6031},
  doi={10.1109/ICCV.2019.00612}
}

@inproceedings{zhong2020random,
  title={Random erasing data augmentation},
  author={Zhong, Zhun and Zheng, Liang and Kang, Guoliang and Li, Shaozi and Yang, Yi},
  booktitle={Proceedings of the AAAI conference on artificial intelligence},
  volume={34},
  number={07},
  pages={13001--13008},
  year={2020}
}

@inproceedings{
zhang2017mixup,
title={mixup: Beyond Empirical Risk Minimization},
author={Hongyi Zhang and Moustapha Cisse and Yann N. Dauphin and David Lopez-Paz},
booktitle={International Conference on Learning Representations},
year={2018},
url={https://openreview.net/forum?id=r1Ddp1-Rb},
}

@article{robbins1951stochastic,
  title={A stochastic approximation method},
  author={Robbins, Herbert and Monro, Sutton},
  journal={The annals of mathematical statistics},
  pages={400--407},
  year={1951},
  publisher={JSTOR},
  doi = {10.1214/aoms/1177729586}
}

@article{amari1993backpropagation,
title = {Backpropagation and stochastic gradient descent method},
journal = {Neurocomputing},
volume = {5},
number = {4},
pages = {185-196},
year = {1993},
issn = {0925-2312},
doi = {https://doi.org/10.1016/0925-2312(93)90006-O},
url = {https://www.sciencedirect.com/science/article/pii/092523129390006O},
author = {Shun-ichi, Amari}
}

@inproceedings{
loshchilov2017decoupled,
title={Decoupled Weight Decay Regularization},
author={Ilya Loshchilov and Frank Hutter},
booktitle={International Conference on Learning Representations},
year={2019},
url={https://openreview.net/forum?id=Bkg6RiCqY7},
}

@article{ballas2015delving,
  title={Delving deeper into convolutional networks for learning video representations},
  author={Ballas, Nicolas and Yao, Li and Pal, Chris and Courville, Aaron},
  journal={arXiv preprint arXiv:1511.06432},
  year={2015}
}

@book{Marshall.2012,
  title={The Cryosphere},
  author={Marshall, Shawn J.},
  year={2012},
  publisher={Princeton University Press},
  place={Princeton, Woodstock},
  isbn = {978-0-691-14525-9}
}

@book{Benn.2014,
  title={Glaciers and glaciation},
  author={Benn, Douglas and Evans, David JA},
  year={2010},
  publisher={Routledge},
  place={New York},
  edition={2},
  isbn = {978-0-340-90579-1}
}

@article{Kochtitzky.2022_2,
 author = {Kochtitzky, William and Copland, Luke and {van Wychen}, Wesley and Hugonnet, Romain and Hock, Regine and Dowdeswell, Julian A. and Benham, Toby and Strozzi, Tazio and Glazovsky, Andrey and Lavrentiev, Ivan and Rounce, David R. and Millan, Romain and Cook, Alison and Dalton, Abigail and Jiskoot, Hester and Cooley, Jade and Jania, Jacek and Navarro, Francisco},
 year = {2022},
 title = {The unquantified mass loss of Northern Hemisphere marine-terminating glaciers from 2000--2020},
 pages = {5835},
 volume = {13},
 number = {1},
 journal = {Nature communications},
 doi = {10.1038/s41467-022-33231-x}
}

@book{IPCC,
   author = {IPCC},
   title = {Climate Change 2021: The Physical Science Basis. Contribution of Working Group I to the Sixth Assessment Report of the Intergovernmental Panel on Climate Change},
   publisher = {Cambridge University Press},
   address = {Cambridge, United Kingdom and New York, NY, USA},
   DOI = {10.1017/9781009157896},
   year = {2021},
   type = {Book}
}

@INPROCEEDINGS{Putatunda.2024,
  author={Putatunda, Rohan and Purushotham, Sanjay and Janeja, Vandana P.},
  booktitle={2024 International Conference on Machine Learning and Applications (ICMLA)}, 
  title={SEATTNET: UNET Enhanced with Squeeze-Excited Attention Gates for Ice-Calving Front Segmentation}, 
  year={2024},
  volume={},
  number={},
  pages={575-582},
  doi={10.1109/ICMLA61862.2024.00084}
}

@INPROCEEDINGS{Maslov.2023,
  author={Maslov, Konstantin A. and Persello, Claudio and Schellenberger, Thomas and Stein, Alfred},
  booktitle={IGARSS 2023 - 2023 IEEE International Geoscience and Remote Sensing Symposium}, 
  title={GLAVITU: A Hybrid CNN-Transformer for Multi-Regional Glacier Mapping from Multi-Source Data}, 
  year={2023},
  volume={},
  number={},
  pages={1233-1236},
  doi={10.1109/IGARSS52108.2023.10281828}
}

@article{Li.2025,
 author = {Li, Tian and Hofer, Stefan and Moholdt, Geir and Igneczi, Adam and Heidler, Konrad and Zhu, Xiao Xiang and Bamber, Jonathan},
 year = {2025},
 title = {Pervasive glacier retreats across Svalbard from 1985 to 2023},
 pages = {705},
 volume = {16},
 number = {1},
 journal = {Nature communications},
 doi = {10.1038/s41467-025-55948-1}
}

@inproceedings{Ronneberger.2015,
 author = {Ronneberger, O. and Fischer, P. and Brox, T.},
 title = {U-Net: Convolutional Networks for Biomedical Image Segmentation},
 pages = {234--241},
 publisher = {{Springer International Publishing}},
 isbn = {978-3-319-24574-4},
 editor = {Navab, Nassir and Hornegger, Joachim and Wells, William M. and Frangi, Alejandro F.},
 booktitle = {Medical Image Computing and Computer-Assisted Intervention (MICCAI)},
 year = {2015},
 address = {Cham},
}

@INPROCEEDINGS{Gourmelon.2023,
  author={Gourmelon, N. and Klink, J. and Seehaus, T. and Braun, M. and Maier, A. and Christlein, V.},
  booktitle={IEEE International Geoscience and Remote Sensing Symposium (IGARSS)},
  title={Conditional Random Fields for improving deep learning-based glacier calving front delineations},
  year={2023},
  volume={},
  number={},
  pages={4939-4942},
  doi={10.1109/IGARSS52108.2023.10282915}
}

@article{Periyasamy.2022,
 author = {Periyasamy, M. and Davari, A. and Seehaus, T. and Braun, M. and Maier, A. and Christlein, V.},
 year = {2022},
 title = {How to Get the Most Out of U-Net for Glacier Calving Front Segmentation},
 volume={15},
 number={},
 pages={1712-1723},
 issn = {1939-1404},
 journal = {IEEE Journal of Selected Topics in Applied Earth Observations and Remote Sensing},
 doi = {10.1109/JSTARS.2022.3148033},
}

@INPROCEEDINGS{Hartmann.2021,
  author={Hartmann, Andreas and Davari, Amirabbas and Seehaus, Thorsten and Braun, Matthias and Maier, Andreas and Christlein, Vincent},
  booktitle={2021 IEEE International Geoscience and Remote Sensing Symposium IGARSS}, 
  title={Bayesian U-Net for Segmenting Glaciers in Sar Imagery}, 
  year={2021},
  volume={},
  number={},
  pages={3479-3482},
  doi={10.1109/IGARSS47720.2021.9554292}
}

@INPROCEEDINGS{Holzmann.2021,
  author={Holzmann, Michael and Davari, Amirabbas and Seehaus, Thorsten and Braun, Matthias and Maier, Andreas and Christlein, Vincent},
  booktitle={2021 IEEE International Geoscience and Remote Sensing Symposium IGARSS}, 
  title={Glacier Calving Front Segmentation Using Attention U-Net}, 
  year={2021},
  volume={},
  number={},
  pages={3483-3486},
  doi={10.1109/IGARSS47720.2021.9555067}
}

@ARTICLE{Davari_Baller.2021,
  author={Davari, A. and Baller, C. and Seehaus, T. and Braun, M. and Maier, A. and Christlein, V.},
  journal={IEEE Transactions on Geoscience and Remote Sensing}, 
  title={Pixelwise Distance Regression for Glacier Calving Front Detection and Segmentation}, 
  year={2022},
  volume={60},
  number={},
  pages={1-10},
  doi={10.1109/TGRS.2022.3158591},
}

@article{Baumhoer.2019,
 author = {Baumhoer, C. A. and Dietz, A. J. and Kneisel, C. and Kuenzer, C.},
 year = {2019},
 title = {Automated Extraction of Antarctic Glacier and Ice Shelf Fronts from Sentinel-1 Imagery Using Deep Learning},
 pages = {2529},
 volume = {11},
 number = {21},
 journal = {Remote Sensing},
 doi = {10.3390/rs11212529}
}

@article{Davari_Islam.2021,
  author={Davari, Amirabbas and Islam, Saahil and Seehaus, Thorsten and Hartmann, Andreas and Braun, Matthias and Maier, Andreas and Christlein, Vincent},
  journal={IEEE Transactions on Geoscience and Remote Sensing}, 
  title={On Mathews Correlation Coefficient and Improved Distance Map Loss for Automatic Glacier Calving Front Segmentation in SAR Imagery}, 
  year={2022},
  volume={60},
  number={},
  pages={1-12},
  doi={10.1109/TGRS.2021.3115883}
}

@article{Cheng.2021,
author = {Cheng, D. and Hayes, W. and Larour, E. and Mohajerani, Y. and Wood, M. and Velicogna, I. and Rignot, E.},
title = {Calving Front Machine (CALFIN): glacial termini dataset  and automated deep learning extraction method  for Greenland, 1972--2019},
journal = {The Cryosphere},
volume = {15},
year = {2021},
number = {3},
pages = {1663--1675},
url = {https://tc.copernicus.org/articles/15/1663/2021/},
doi = {10.5194/tc-15-1663-2021}
}

@article{Heidler.2021,
 author = {Heidler, K. and Mou, L. and Baumhoer, C. and Dietz, A. and Zhu, X. X.},
 year = {2021},
 title = {HED-UNet: Combined Segmentation and Edge Detection for Monitoring the Antarctic Coastline},
 pages = {1--14},
 volume={60},
 journal = {IEEE Transactions on Geoscience and Remote Sensing},
 doi = {10.1109/TGRS.2021.3064606}
}

@Article{Herrmann.2023,
AUTHOR = {Herrmann, O. and Gourmelon, N. and Seehaus, T. and Maier, A. and F\"urst, J. J. and Braun, M. H. and Christlein, V.},
TITLE = {Out-of-the-box calving-front detection method using deep learning},
JOURNAL = {The Cryosphere},
VOLUME = {17},
YEAR = {2023},
NUMBER = {11},
PAGES = {4957--4977},
URL = {https://tc.copernicus.org/articles/17/4957/2023/},
DOI = {10.5194/tc-17-4957-2023}
}

@ARTICLE{Loebel.2022,
  author={Loebel, Erik and Scheinert, Mirko and Horwath, Martin and Heidler, Konrad and Christmann, Julia and Phan, Long Duc and Humbert, Angelika and Zhu, Xiao Xiang},
  journal={IEEE Transactions on Geoscience and Remote Sensing}, 
  title={Extracting Glacier Calving Fronts by Deep Learning: The Benefit of Multispectral, Topographic, and Textural Input Features}, 
  year={2022},
  volume={60},
  number={},
  pages={1-12},
  doi={10.1109/TGRS.2022.3208454}}

@article{Mohajerani.2019,
 author = {Mohajerani, Y. and Wood, M. and Velicogna, I. and Rignot, E.},
 year = {2019},
 title = {Detection of Glacier Calving Margins with Convolutional Neural Networks: A Case Study},
 pages = {74},
 volume = {11},
 number = {1},
 journal = {Remote Sensing},
 doi = {10.3390/rs11010074 }
}

@ARTICLE{Wu.2023_1,
  author={Wu, Fei and Gourmelon, Nora and Seehaus, Thorsten and Zhang, Jianlin and Braun, Matthias and Maier, Andreas and Christlein, Vincent},
  journal={IEEE Transactions on Geoscience and Remote Sensing}, 
  title={AMD-HookNet for Glacier Front Segmentation}, 
  year={2023},
  volume={61},
  number={},
  pages={1-12},
  doi={10.1109/TGRS.2023.3245419}}

@ARTICLE{Wu.2023_2,
  author={Wu, Fei and Gourmelon, Nora and Seehaus, Thorsten and Zhang, Jianlin and Braun, Matthias and Maier, Andreas and Christlein, Vincent},
  journal={IEEE Transactions on Geoscience and Remote Sensing}, 
  title={Contextual HookFormer for Glacier Calving Front Segmentation}, 
  year={2024},
  volume={62},
  number={},
  pages={1-15},
  doi={10.1109/TGRS.2024.3368215}}

@article{Zhang.2019,
 author = {Zhang, E. and Liu, L. and Huang, L.},
 year = {2019},
 title = {Automatically delineating the calving front of Jakobshavn Isbr{\ae} from multitemporal TerraSAR-X images: a deep learning approach},
 pages = {1729--1741},
 volume = {13},
 number = {6},
 journal = {The Cryosphere},
 doi = {10.5194/tc-13-1729-2019},
}

@article{Zhang.2021                           ,
 author = {Zhang, E. and Liu, L. and Huang, L. and Ng, K. S.},
 year = {2021},
 title = {An automated, generalized, deep-learning-based method for delineating the calving fronts of Greenland glaciers from multi-sensor remote sensing imagery},
 pages = {112265},
 volume = {254},
 issn = {00344257},
 journal = {Remote Sensing of Environment},
 doi = {10.1016/j.rse.2020.112265}
}

@article{Zhang.2023,
 author = {Zhang, E. and Catania, G. and Trugman, D. T.},
 year = {2023},
 title = {AutoTerm: an automated pipeline for glacier terminus extraction using machine learning and a ``big data'' repository of Greenland glacier termini},
 url = {https://tc.copernicus.org/articles/17/3485/2023/},
 pages = {3485--3503},
 volume = {17},
 number = {8},
 journal = {The Cryosphere},
 doi = {10.5194/tc-17-3485-2023},
}

@ARTICLE{Zhu.2023,
  author={Zhu, Qi and Guo, Huadong and Zhang, Lu and Liang, Dong and Wu, Zherong and Liu, Yiming and Lv, Zhuoran},
  journal={IEEE Transactions on Geoscience and Remote Sensing}, 
  title={GLA-STDeepLab: SAR Enhancing Glacier and Ice Shelf Front Detection Using Swin-TransDeepLab With Global–Local Attention}, 
  year={2023},
  volume={61},
  number={},
  pages={1-13},
  doi={10.1109/TGRS.2023.3324404}}

@article{xie2021segformer,
  title={SegFormer: Simple and efficient design for semantic segmentation with transformers},
  author={Xie, Enze and Wang, Wenhai and Yu, Zhiding and Anandkumar, Anima and Alvarez, Jose M and Luo, Ping},
  journal={Advances in neural information processing systems},
  volume={34},
  pages={12077--12090},
  year={2021}
}

@article{chen2019rethinking,
  title={Rethinking atrous convolution for semantic image segmentation. arXiv 2017},
  author={Chen, Liang-Chieh and Papandreou, George and Schroff, Florian and Adam, Hartwig},
  journal={arXiv preprint arXiv:1706.05587},
  volume={2},
  pages={1},
  year={2019}
}

@INPROCEEDINGS {jain2023semask,
author = { Jain, Jitesh and Singh, Anukriti and Orlov, Nikita and Huang, Zilong and Li, Jiachen and Walton, Steven and Shi, Humphrey },
booktitle = { 2023 IEEE/CVF International Conference on Computer Vision Workshops (ICCVW) },
title = {{ SeMask: Semantically Masked Transformers for Semantic Segmentation }},
year = {2023},
volume = {},
ISSN = {},
pages = {752-761},
doi = {10.1109/ICCVW60793.2023.00083},
url = {https://doi.ieeecomputersociety.org/10.1109/ICCVW60793.2023.00083},
publisher = {IEEE Computer Society},
address = {Los Alamitos, CA, USA},
month =Oct}

@misc{ramachandran2017searching,
title={Searching for Activation Functions},
author={Prajit Ramachandran, Barret Zoph, Quoc V. Le},
year={2018},
url={https://openreview.net/forum?id=SkBYYyZRZ},
}

@inproceedings{wu2018group,
  title={Group normalization},
  author={Wu, Yuxin and He, Kaiming},
  booktitle={Proceedings of the European conference on computer vision (ECCV)},
  pages={3--19},
  year={2018}
}

@INPROCEEDINGS{zhang2023adding,
  author={Zhang, Lvmin and Rao, Anyi and Agrawala, Maneesh},
  booktitle={2023 IEEE/CVF International Conference on Computer Vision (ICCV)}, 
  title={Adding Conditional Control to Text-to-Image Diffusion Models}, 
  year={2023},
  volume={},
  number={},
  pages={3813-3824},
  doi={10.1109/ICCV51070.2023.00355}
}

@inproceedings{bachlechner2021rezero,
  title={Rezero is all you need: Fast convergence at large depth},
  author={Bachlechner, Thomas and Majumder, Bodhisattwa Prasad and Mao, Henry and Cottrell, Gary and McAuley, Julian},
  booktitle={Uncertainty in Artificial Intelligence},
  pages={1352--1361},
  year={2021},
  organization={PMLR}
}

@article{he2024time,
title = {Time-series land cover change detection using deep learning-based temporal semantic segmentation},
journal = {Remote Sensing of Environment},
volume = {305},
pages = {114101},
year = {2024},
issn = {0034-4257},
doi = {https://doi.org/10.1016/j.rse.2024.114101},
url = {https://www.sciencedirect.com/science/article/pii/S0034425724001123},
author = {Haixu He and Jining Yan and Dong Liang and Zhongchang Sun and Jun Li and Lizhe Wang}
}

@ARTICLE{saha2020unsupervised,
  author={Saha, Sudipan and Mou, Lichao and Qiu, Chunping and Zhu, Xiao Xiang and Bovolo, Francesca and Bruzzone, Lorenzo},
  journal={IEEE Transactions on Geoscience and Remote Sensing}, 
  title={Unsupervised Deep Joint Segmentation of Multitemporal High-Resolution Images}, 
  year={2020},
  volume={58},
  number={12},
  pages={8780-8792},
  doi={10.1109/TGRS.2020.2990640}}

@inproceedings{tarasiou2023vits,
  title={Vits for sits: Vision transformers for satellite image time series},
  author={Tarasiou, Michail and Chavez, Erik and Zafeiriou, Stefanos},
  booktitle={Proceedings of the IEEE/CVF Conference on Computer Vision and Pattern Recognition},
  pages={10418--10428},
  year={2023}
}

@INPROCEEDINGS{garnot2021panoptic,
  author={Fare Garnot, Vivien Sainte and Landrieu, Loic},
  booktitle={2021 IEEE/CVF International Conference on Computer Vision (ICCV)}, 
  title={Panoptic Segmentation of Satellite Image Time Series with Convolutional Temporal Attention Networks}, 
  year={2021},
  volume={},
  number={},
  pages={4852-4861},
  doi={10.1109/ICCV48922.2021.00483}
}

@INPROCEEDINGS{garnot2020satellite,
  author={Sainte Fare Garnot, Vivien and Landrieu, Loic and Giordano, Sebastien and Chehata, Nesrine},
  booktitle={2020 IEEE/CVF Conference on Computer Vision and Pattern Recognition (CVPR)}, 
  title={Satellite Image Time Series Classification With Pixel-Set Encoders and Temporal Self-Attention}, 
  year={2020},
  volume={},
  number={},
  pages={12322-12331},
  doi={10.1109/CVPR42600.2020.01234}
}

@INPROCEEDINGS{lai2021semi,
  author={Lai, Xin and Tian, Zhuotao and Jiang, Li and Liu, Shu and Zhao, Hengshuang and Wang, Liwei and Jia, Jiaya},
  booktitle={2021 IEEE/CVF Conference on Computer Vision and Pattern Recognition (CVPR)}, 
  title={Semi-supervised Semantic Segmentation with Directional Context-aware Consistency}, 
  year={2021},
  volume={},
  number={},
  pages={1205-1214},
  doi={10.1109/CVPR46437.2021.00126}}

@INPROCEEDINGS{liu2021swin,
author = { Liu, Ze and Lin, Yutong and Cao, Yue and Hu, Han and Wei, Yixuan and Zhang, Zheng and Lin, Stephen and Guo, Baining },
booktitle = { 2021 IEEE/CVF International Conference on Computer Vision (ICCV) },
title = {{ Swin Transformer: Hierarchical Vision Transformer using Shifted Windows }},
year = {2021},
volume = {},
ISSN = {},
pages = {9992-10002},
doi = {10.1109/ICCV48922.2021.00986},
url = {https://doi.ieeecomputersociety.org/10.1109/ICCV48922.2021.00986},
publisher = {IEEE Computer Society},
address = {Los Alamitos, CA, USA},
month =Oct}

@INPROCEEDINGS{strudel2021segmenter,
  author={Strudel, Robin and Garcia, Ricardo and Laptev, Ivan and Schmid, Cordelia},
  booktitle={2021 IEEE/CVF International Conference on Computer Vision (ICCV)}, 
  title={Segmenter: Transformer for Semantic Segmentation}, 
  year={2021},
  volume={},
  number={},
  pages={7242-7252},
  doi={10.1109/ICCV48922.2021.00717}
}

@inproceedings{
dosovitskiy2020image,
title={An Image is Worth 16x16 Words: Transformers for Image Recognition at Scale},
author={Alexey Dosovitskiy and Lucas Beyer and Alexander Kolesnikov and Dirk Weissenborn and Xiaohua Zhai and Thomas Unterthiner and Mostafa Dehghani and Matthias Minderer and Georg Heigold and Sylvain Gelly and Jakob Uszkoreit and Neil Houlsby},
booktitle={International Conference on Learning Representations},
year={2021},
url={https://openreview.net/forum?id=YicbFdNTTy}
}

@Article{feng2023smbcnet,
AUTHOR = {Feng, Jiangfan and Yang, Xinyu and Gu, Zhujun and Zeng, Maimai and Zheng, Wei},
TITLE = {SMBCNet: A Transformer-Based Approach for Change Detection in Remote Sensing Images through Semantic Segmentation},
JOURNAL = {Remote Sensing},
VOLUME = {15},
YEAR = {2023},
NUMBER = {14},
ARTICLE-NUMBER = {3566},
URL = {https://www.mdpi.com/2072-4292/15/14/3566},
ISSN = {2072-4292},
DOI = {10.3390/rs15143566}
}

@article{marsocci2023inferring,
title = {Inferring 3D change detection from bitemporal optical images},
journal = {ISPRS Journal of Photogrammetry and Remote Sensing},
volume = {196},
pages = {325-339},
year = {2023},
issn = {0924-2716},
doi = {https://doi.org/10.1016/j.isprsjprs.2022.12.009},
url = {https://www.sciencedirect.com/science/article/pii/S0924271622003240},
author = {Valerio Marsocci and Virginia Coletta and Roberta Ravanelli and Simone Scardapane and Mattia Crespi}
}

@ARTICLE{fang2023changer,
  author={Fang, Sheng and Li, Kaiyu and Li, Zhe},
  journal={IEEE Transactions on Geoscience and Remote Sensing}, 
  title={Changer: Feature Interaction is What You Need for Change Detection}, 
  year={2023},
  volume={61},
  number={},
  pages={1-11},
  doi={10.1109/TGRS.2023.3277496}
}

@ARTICLE{li2023lightweight,
  author={Li, Zhenglai and Tang, Chang and Liu, Xinwang and Zhang, Wei and Dou, Jie and Wang, Lizhe and Zomaya, Albert Y.},
  journal={IEEE Transactions on Geoscience and Remote Sensing}, 
  title={Lightweight Remote Sensing Change Detection With Progressive Feature Aggregation and Supervised Attention}, 
  year={2023},
  volume={61},
  number={},
  pages={1-12},
  doi={10.1109/TGRS.2023.3241436}
}

@INPROCEEDINGS{bernhard2023mapformer,
  author={Bernhard, Maximilian and Strauß, Niklas and Schubert, Matthias},
  booktitle={2023 IEEE/CVF International Conference on Computer Vision (ICCV)}, 
  title={MapFormer: Boosting Change Detection by Using Pre-change Information}, 
  year={2023},
  volume={},
  number={},
  pages={16791-16800},
  doi={10.1109/ICCV51070.2023.01544}
}

@article{zheng2022changemask,
title = {ChangeMask: Deep multi-task encoder-transformer-decoder architecture for semantic change detection},
journal = {ISPRS Journal of Photogrammetry and Remote Sensing},
volume = {183},
pages = {228-239},
year = {2022},
issn = {0924-2716},
doi = {https://doi.org/10.1016/j.isprsjprs.2021.10.015},
url = {https://www.sciencedirect.com/science/article/pii/S0924271621002835},
author = {Zhuo Zheng and Yanfei Zhong and Shiqi Tian and Ailong Ma and Liangpei Zhang}
}

@article{cui2023mtscd,
title = {MTSCD-Net: A network based on multi-task learning for semantic change detection of bitemporal remote sensing images},
journal = {International Journal of Applied Earth Observation and Geoinformation},
volume = {118},
pages = {103294},
year = {2023},
issn = {1569-8432},
doi = {https://doi.org/10.1016/j.jag.2023.103294},
url = {https://www.sciencedirect.com/science/article/pii/S1569843223001164},
author = {Fengzhi Cui and Jie Jiang}
}

@article{daudt2019multitask,
title = {Multitask learning for large-scale semantic change detection},
journal = {Computer Vision and Image Understanding},
volume = {187},
pages = {102783},
year = {2019},
issn = {1077-3142},
doi = {https://doi.org/10.1016/j.cviu.2019.07.003},
url = {https://www.sciencedirect.com/science/article/pii/S1077314219300992},
author = {Rodrigo {Caye Daudt} and Bertrand {Le Saux} and Alexandre Boulch and Yann Gousseau}
}

@article{ding2022bi,
  title={Bi-temporal semantic reasoning for the semantic change detection in HR remote sensing images},
  author={Ding, Lei and Guo, Haitao and Liu, Sicong and Mou, Lichao and Zhang, Jing and Bruzzone, Lorenzo},
  journal={IEEE Transactions on Geoscience and Remote Sensing},
  volume={60},
  pages={1--14},
  year={2022},
  publisher={IEEE}
}

@ARTICLE{ding2024joint,
  author={Ding, Lei and Zhang, Jing and Guo, Haitao and Zhang, Kai and Liu, Bing and Bruzzone, Lorenzo},
  journal={IEEE Transactions on Geoscience and Remote Sensing}, 
  title={Joint Spatio-Temporal Modeling for Semantic Change Detection in Remote Sensing Images}, 
  year={2024},
  volume={62},
  number={},
  pages={1-14},
  doi={10.1109/TGRS.2024.3362795}
}

@Article{jiang2023ttnet,
AUTHOR = {Jiang, Liangcun and Li, Feng and Huang, Li and Peng, Feifei and Hu, Lei},
TITLE = {TTNet: A Temporal-Transform Network for Semantic Change Detection Based on Bi-Temporal Remote Sensing Images},
JOURNAL = {Remote Sensing},
VOLUME = {15},
YEAR = {2023},
NUMBER = {18},
ARTICLE-NUMBER = {4555},
URL = {https://www.mdpi.com/2072-4292/15/18/4555},
ISSN = {2072-4292},
DOI = {10.3390/rs15184555}
}

@ARTICLE{liu2024tbscd,
  author={Liu, Xuanguang and Dai, Chenguang and Zhang, Zhenchao and Li, Mengmeng and Wang, Hanyun and Ji, Hongliang and Li, Yujie},
  journal={IEEE Geoscience and Remote Sensing Letters}, 
  title={TBSCD-Net: A Siamese Multitask Network Integrating Transformers and Boundary Regularization for Semantic Change Detection From VHR Satellite Images}, 
  year={2024},
  volume={21},
  number={},
  pages={1-5},
  doi={10.1109/LGRS.2024.3385404}
}

@article{tian2022large,
title = {Large-scale deep learning based binary and semantic change detection in ultra high resolution remote sensing imagery: From benchmark datasets to urban application},
journal = {ISPRS Journal of Photogrammetry and Remote Sensing},
volume = {193},
pages = {164-186},
year = {2022},
issn = {0924-2716},
doi = {https://doi.org/10.1016/j.isprsjprs.2022.08.012},
url = {https://www.sciencedirect.com/science/article/pii/S0924271622002210},
author = {Shiqi Tian and Yanfei Zhong and Zhuo Zheng and Ailong Ma and Xicheng Tan and Liangpei Zhang}
}

@ARTICLE{bruzzone2002iterative,
  author={Bruzzone, L. and Serpico, S.B.},
  journal={IEEE Transactions on Geoscience and Remote Sensing}, 
  title={An iterative technique for the detection of land-cover transitions in multitemporal remote-sensing images}, 
  year={1997},
  volume={35},
  number={4},
  pages={858-867},
  doi={10.1109/36.602528}
}

@article{weismiller1977change,
  title={Change detection in coastal zone environments},
  author={Weismiller, RA and Kristof, SJ and Scholz, DK and Anuta, PE and Momin, SA},
  journal={Photogrammetric Engineering and Remote Sensing},
  volume={43},
  number={12},
  pages={1533--1539},
  year={1977}
}

@ARTICLE{xia2022deep,
  author={Xia, Hao and Tian, Yugang and Zhang, Lihao and Li, Shuangliang},
  journal={IEEE Transactions on Geoscience and Remote Sensing}, 
  title={A Deep Siamese Postclassification Fusion Network for Semantic Change Detection}, 
  year={2022},
  volume={60},
  number={},
  pages={1-16},
  doi={10.1109/TGRS.2022.3171067}
}

@ARTICLE{yang2020semantic,
  author={Yang, Kunping and Xia, Gui-Song and Liu, Zicheng and Du, Bo and Yang, Wen and Pelillo, Marcello and Zhang, Liangpei},
  journal={IEEE Transactions on Geoscience and Remote Sensing}, 
  title={Asymmetric Siamese Networks for Semantic Change Detection in Aerial Images}, 
  year={2022},
  volume={60},
  number={},
  pages={1-18},
  doi={10.1109/TGRS.2021.3113912}
}

@article{yuan2022transformer,
  title={A transformer-based Siamese network and an open optical dataset for semantic change detection of remote sensing images},
  author={Yuan, Panli and Zhao, Qingzhan and Zhao, Xingbiao and Wang, Xuewen and Long, Xuefeng and Zheng, Yuchen},
  journal={International Journal of Digital Earth},
  volume={15},
  number={1},
  pages={1506--1525},
  year={2022},
  publisher={Taylor \& Francis},
  doi = {10.1080/17538947.2022.2111470}
}

@ARTICLE{Zhao2022-ta,
  author={Zhao, Manqi and Zhao, Zifei and Gong, Shuai and Liu, Yunfei and Yang, Jian and Xiong, Xiong and Li, Shengyang},
  journal={IEEE Journal of Selected Topics in Applied Earth Observations and Remote Sensing}, 
  title={Spatially and Semantically Enhanced Siamese Network for Semantic Change Detection in High-Resolution Remote Sensing Images}, 
  year={2022},
  volume={15},
  number={},
  pages={2563-2573},
  doi={10.1109/JSTARS.2022.3159528}
}

@article{Zheng2022-oz,
title = {ChangeMask: Deep multi-task encoder-transformer-decoder architecture for semantic change detection},
journal = {ISPRS Journal of Photogrammetry and Remote Sensing},
volume = {183},
pages = {228-239},
year = {2022},
issn = {0924-2716},
doi = {https://doi.org/10.1016/j.isprsjprs.2021.10.015},
url = {https://www.sciencedirect.com/science/article/pii/S0924271621002835},
author = {Zhuo Zheng and Yanfei Zhong and Shiqi Tian and Ailong Ma and Liangpei Zhang}
}

@INPROCEEDINGS{papadomanolaki2019detecting,
  author={Papadomanolaki, Maria and Verma, Sagar and Vakalopoulou, Maria and Gupta, Siddharth and Karantzalos, Konstantinos},
  booktitle={IGARSS 2019 - 2019 IEEE International Geoscience and Remote Sensing Symposium}, 
  title={Detecting Urban Changes with Recurrent Neural Networks from Multitemporal Sentinel-2 Data}, 
  year={2019},
  volume={},
  number={},
  pages={214-217},
  doi={10.1109/IGARSS.2019.8900330}
}

@ARTICLE{hafner2024continuous,
  author={Hafner, Sebastian and Fang, Heng and Azizpour, Hossein and Ban, Yifang},
  journal={IEEE Transactions on Geoscience and Remote Sensing}, 
  title={Continuous Urban Change Detection From Satellite Image Time Series With Temporal Feature Refinement and Multitask Integration}, 
  year={2025},
  volume={63},
  number={},
  pages={1-18},
  doi={10.1109/TGRS.2025.3578866}
}

@ARTICLE{chen2024unsupervised,
  author={Chen, Yuxing and Bruzzone, Lorenzo},
  journal={IEEE Transactions on Geoscience and Remote Sensing}, 
  title={Unsupervised CD in Satellite Image Time Series by Contrastive Learning and Feature Tracking}, 
  year={2024},
  volume={62},
  number={},
  pages={1-13},
  doi={10.1109/TGRS.2024.3354118}}

@inproceedings{esser2021taming,
  title={Taming transformers for high-resolution image synthesis},
  author={Esser, Patrick and Rombach, Robin and Ommer, Bjorn},
  booktitle={Proceedings of the IEEE/CVF conference on computer vision and pattern recognition},
  pages={12873--12883},
  year={2021}
}

@Article{cheng2023application,
AUTHOR = {Cheng, Xinglu and Sun, Yonghua and Zhang, Wangkuan and Wang, Yihan and Cao, Xuyue and Wang, Yanzhao},
TITLE = {Application of Deep Learning in Multitemporal Remote Sensing Image Classification},
JOURNAL = {Remote Sensing},
VOLUME = {15},
YEAR = {2023},
NUMBER = {15},
ARTICLE-NUMBER = {3859},
URL = {https://www.mdpi.com/2072-4292/15/15/3859},
ISSN = {2072-4292},
DOI = {10.3390/rs15153859}
}

@Article{zhao2025histenet,
AUTHOR = {Zhao, Lu and Wan, Ling and Ma, Lei and Zhang, Yiming},
TITLE = {HiSTENet: History-Integrated Spatial–Temporal Information Extraction Network for Time Series Remote Sensing Image Change Detection},
JOURNAL = {Remote Sensing},
VOLUME = {17},
YEAR = {2025},
NUMBER = {5},
ARTICLE-NUMBER = {792},
URL = {https://www.mdpi.com/2072-4292/17/5/792},
ISSN = {2072-4292},
DOI = {10.3390/rs17050792}
}

@Article{russwurm2018multi,
AUTHOR = {Rußwurm, Marc and Körner, Marco},
TITLE = {Multi-Temporal Land Cover Classification with Sequential Recurrent Encoders},
JOURNAL = {ISPRS International Journal of Geo-Information},
VOLUME = {7},
YEAR = {2018},
NUMBER = {4},
ARTICLE-NUMBER = {129},
URL = {https://www.mdpi.com/2220-9964/7/4/129},
ISSN = {2220-9964},
DOI = {10.3390/ijgi7040129}
}

@Article{yang2022multi,
AUTHOR = {Yang, Xuan and Zhang, Bing and Chen, Zhengchao and Bai, Yongqing and Chen, Pan},
TITLE = {A Multi-Temporal Network for Improving Semantic Segmentation of Large-Scale Landsat Imagery},
JOURNAL = {Remote Sensing},
VOLUME = {14},
YEAR = {2022},
NUMBER = {19},
ARTICLE-NUMBER = {5062},
URL = {https://www.mdpi.com/2072-4292/14/19/5062},
ISSN = {2072-4292},
DOI = {10.3390/rs14195062}
}

@article{arfa2024utilizing,
title = {Utilizing multitemporal indices and spectral bands of Sentinel-2 to enhance land use and land cover classification with random forest and support vector machine},
journal = {Advances in Space Research},
volume = {74},
number = {11},
pages = {5580-5590},
year = {2024},
issn = {0273-1177},
doi = {https://doi.org/10.1016/j.asr.2024.08.062},
url ={https://www.sciencedirect.com/science/article/pii/S027311772400886X},
author = {Atefe, Arfa and Masoud, Minaei}
}

@ARTICLE{schuster1997bidirectional,
  author={Schuster, M. and Paliwal, K.K.},
  journal={IEEE Transactions on Signal Processing}, 
  title={Bidirectional recurrent neural networks}, 
  year={1997},
  volume={45},
  number={11},
  pages={2673-2681},
  doi={10.1109/78.650093}
}

@INPROCEEDINGS{maslov2024glacier,
  author={Maslov, Konstantin A. and Schellenberger, Thomas and Persello, Claudio and Stein, Alfred},
  booktitle={IGARSS 2024 - 2024 IEEE International Geoscience and Remote Sensing Symposium}, 
  title={Glacier Mapping from Sentinel-1 SAR Time Series with Deep Learning in Svalbard}, 
  year={2024},
  volume={},
  number={},
  pages={14-17},
  doi={10.1109/IGARSS53475.2024.10640676}
}

@ARTICLE{tyrion,
  author={Gourmelon, Nora and Dreier, Marcel and Mayr, Martin and Seehaus, Thorsten and Pyles, Dakota and Braun, Matthias and Maier, Andreas and Christlein, Vincent},
  journal={IEEE Transactions on Geoscience and Remote Sensing}, 
  title={SSL4SAR: Self-Supervised Learning for Glacier Calving Front Extraction From SAR Imagery}, 
  year={2025},
  volume={63},
  number={},
  pages={1-12},
  doi={10.1109/TGRS.2025.3580945}
}

@article{voelsen2024transformer,
  title={Transformer models for land cover classification with satellite image time series},
  author={Voelsen, Mirjana and Rottensteiner, Franz and Heipke, Christian},
  journal={PFG--Journal of Photogrammetry, Remote Sensing and Geoinformation Science},
  volume={92},
  number={5},
  pages={547--568},
  year={2024},
  publisher={Springer},
  doi={10.3390/rs15071860}
}

@article{ZHAO2025666,
title = {CISNet: Change information guided semantic segmentation network for automatic extraction of glacier calving fronts},
journal = {ISPRS Journal of Photogrammetry and Remote Sensing},
volume = {228},
pages = {666-678},
year = {2025},
issn = {0924-2716},
doi = {https://doi.org/10.1016/j.isprsjprs.2025.08.001},
url = {https://www.sciencedirect.com/science/article/pii/S0924271625003120},
author = {Ji Zhao and Jiayu Tong and Tianhong Li and Yao Sun and Changliang Shao and Yuting Dong}
}

\end{document}